\documentclass{article}

\usepackage{microtype}
\usepackage{graphicx}
\usepackage{tikz}
\usepackage{subcaption}
\usepackage{booktabs} 
\usepackage{xspace}

\usepackage{hyperref}
\usepackage{caption}
\usepackage{multirow}
\usepackage{pgfplots}
\usepackage{overpic}

\usepackage[preprint]{icml2026}

\usepackage{amsmath}
\usepackage{amssymb}
\usepackage{mathtools}
\usepackage{amsthm}
\usepackage{iac_pkg}
\usepackage{subcaption} 

\DeclareMathOperator*{\argmin}{arg\,min}

\newcommand{\modelname}{\textbf{\texttt{HyCon}}\xspace}

\newcommand{\icmlEqualSupervision}{\textsuperscript{$\dagger$}Equal supervision. }
\usepackage{amsmath,amssymb}

\newcommand{\bv}{\mathbf{v}}

\usepackage[capitalize,noabbrev]{cleveref}

\theoremstyle{plain}

\theoremstyle{definition}

\theoremstyle{remark}

\usepackage[textsize=tiny]{todonotes}

\icmltitlerunning{Not All Latent Spaces Are Flat: Hyperbolic Concept Control}

\begin{document}
\definecolor{lightred}{rgb}{0.8,0.1,0}
\twocolumn[
\icmltitle{Not All Latent Spaces Are Flat: Hyperbolic Concept Control}



  \icmlsetsymbol{equal}{*}
  \icmlsetsymbol{super}{$\dagger$}

  \begin{icmlauthorlist}
    \icmlauthor{Maria Rosaria Briglia}{sap}
    \icmlauthor{Simone Facchiano}{sap}
    \icmlauthor{Paolo Cursi}{sap}
    \icmlauthor{Alessio Sampieri}{ital}
    \icmlauthor{Emanuele Rodolà}{sap,par}
    \icmlauthor{Guido Maria D'Amely di Melendugno}{sap}
    \icmlauthor{Luca Franco}{ital}
    \icmlauthor{Fabio Galasso}{sap,super}
    \icmlauthor{Iacopo Masi}{sap,super}
  \end{icmlauthorlist}

  \icmlaffiliation{sap}{Department of Computer Science, Sapienza University of Rome, Rome, Italy}
  \icmlaffiliation{ital}{ItalAI Labs, Rome, Italy}
  \icmlaffiliation{par}{Paradigma Inc., Rome, Italy}

  \icmlcorrespondingauthor{Maria Rosaria Briglia}{briglia@di.uniroma1.it}
  \icmlcorrespondingauthor{Simone Facchiano}{simone.facchiano@uniroma1.it}

  \icmlkeywords{Machine Learning, ICML}

]
\vspace{0.2pt}



\printAffiliationsAndNotice{\icmlEqualSupervision}  

\begin{abstract}

As modern text-to-image (T2I) models draw closer to synthesizing highly realistic content, the threat of unsafe content generation grows, and it becomes paramount to exercise control.
Existing approaches steer these models by applying Euclidean adjustments to text embeddings, redirecting the generation away from unsafe concepts.
In this work, we introduce hyperbolic control (\modelname): a novel control mechanism based on parallel transport that leverages semantically aligned hyperbolic representation space to yield more expressive and stable manipulation of concepts. 
\modelname reuses off-the-shelf generative models and a state-of-the-art hyperbolic text encoder, linked via a lightweight adapter.
\modelname achieves state-of-the-art results across four safety benchmarks and four T2I backbones, showing that hyperbolic steering is a practical and flexible approach for more reliable T2I generation. 
    
\end{abstract}

\section{Introduction}

\begin{figure}[t]
    \centering
    \begin{overpic}[width=.95\columnwidth]{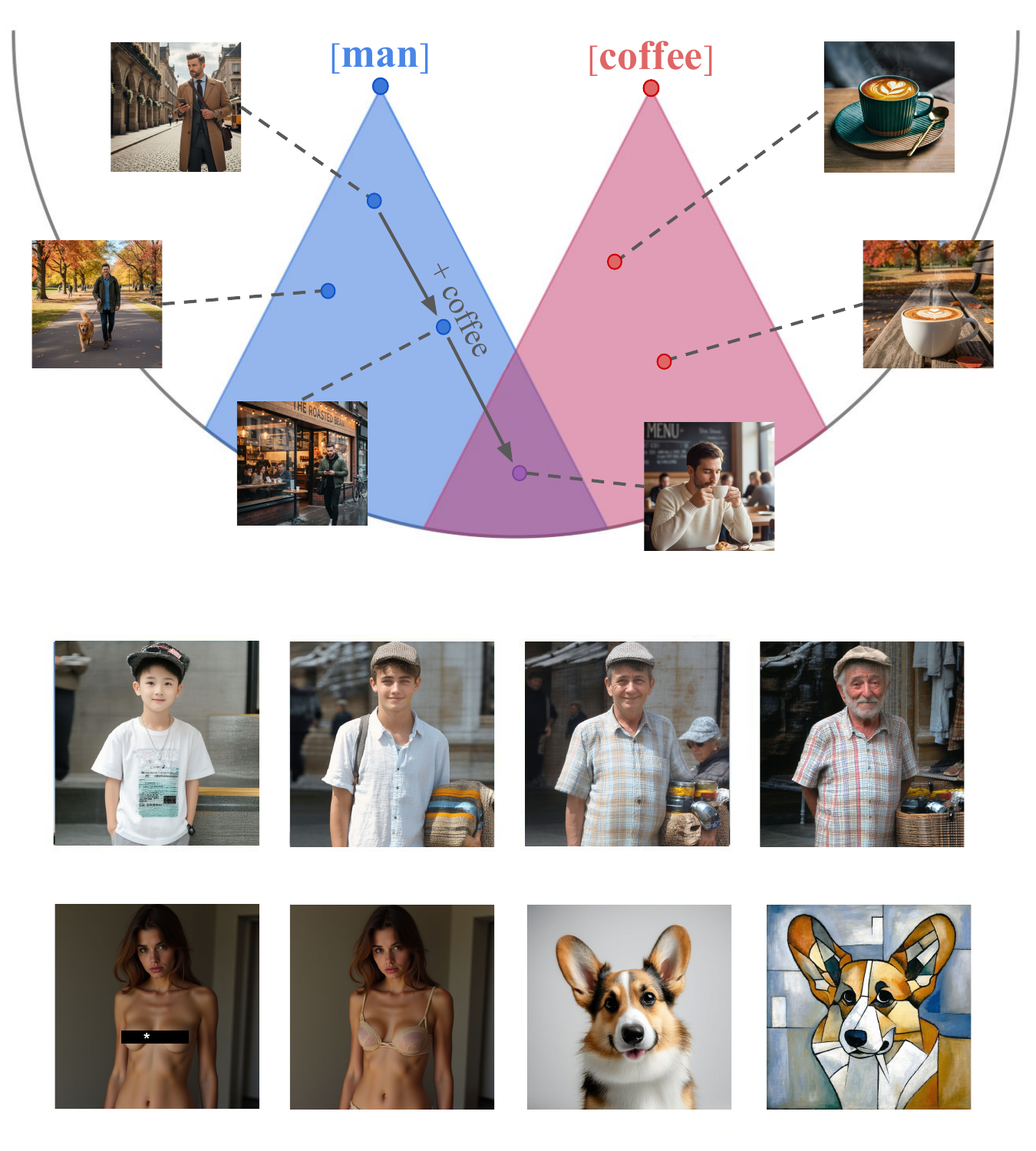}
        \put(15,46){\small Interpolate \texttt{`Young'} to \texttt{`Elder'}}
        \put(9,22.8){\footnotesize Remove \texttt{`Nudity'}}
        \put(51,22.8){\footnotesize Add \texttt{`Picasso'}}
    \end{overpic}
    \caption{(\textit{top}) In hyperbolic space, concepts (e.g., man or coffee) form entailment cones, and concepts' composition corresponds to the cones' intersection. To edit a prompt embedding (e.g., adding coffee to man), we steer it toward the corresponding intersection. (\textit{bottom}) \modelname{} leverages this hyperbolic geometric structure to add or remove concepts via geometry-consistent edits. 
    }
    \label{fig:teaser}
    \vspace{-.45cm}
\end{figure}

The rapid progress of generative models, particularly diffusion models, has enabled high-quality image synthesis from natural language prompts, substantially lowering the barrier to visual content creation.
Beyond prompt engineering, a growing line of work has explored \emph{concept control by steering}, i.e., post-hoc manipulation of model representations to encourage or suppress specific semantic attributes during generation~\cite{dathathriplug,leeprogramming,schramowski2023safe,yoon2025safree}.
Most existing steering approaches operate in Euclidean embedding spaces inherited from vision--language models (VLMs) such as CLIP~\cite{Radford2021}, and rely on linear manipulations of text or latent representations, including inference-time vector steering~\cite{schramowski2023safe,yoon2025safree}, direct weight editing~\cite{gandikota2023erasing,li2024rece}, or activation-level modulation~\cite{zhang2024ant}.
While these methods provide flexible control without retraining, they offer limited structural guarantees on how semantic changes evolve, making their behavior difficult to predict and regulate under substantial interventions.

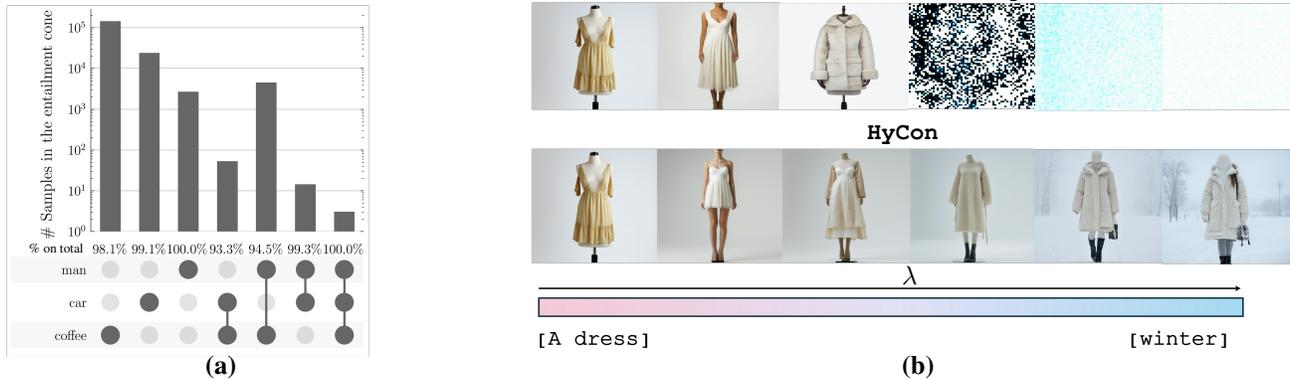
\begin{figure*}[t]
    \centering
    \vspace{8mm}
    \begin{overpic}[width=0.9\linewidth, trim=60 5 65 45]{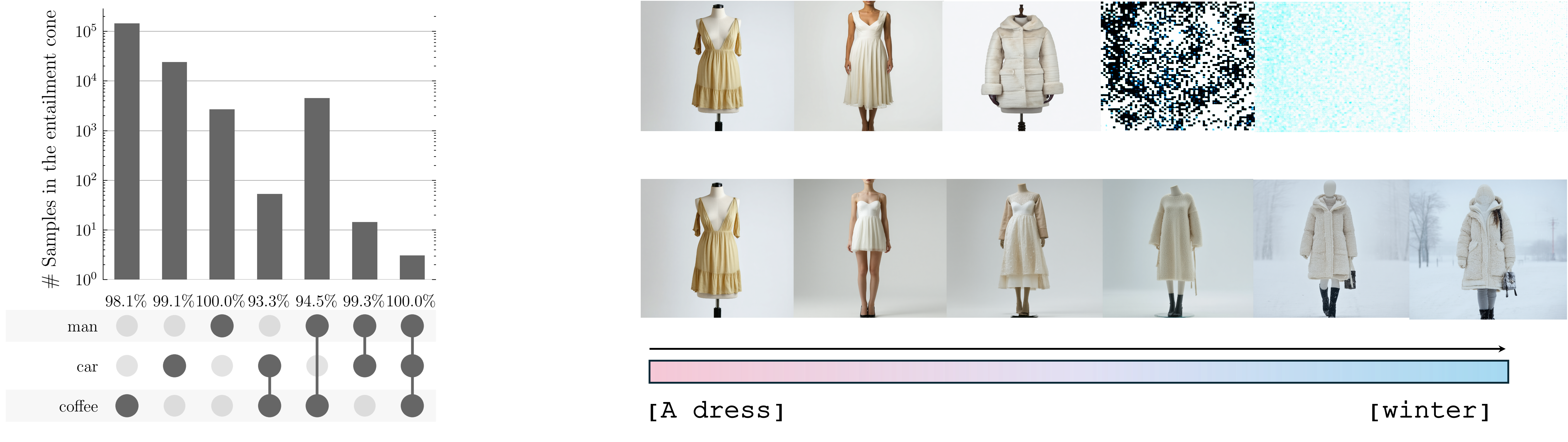}
        \put(12, -1.2){\textbf{(a)}}
        \put(72, -1.2){\textbf{(b)}}
        \put(-3.2,9.2){\fontsize{5pt}{6pt}\selectfont \% on total}

        \put(67,31.3){\footnotesize Euclidean steering}
        \put(69,19){\footnotesize \modelname{}}
        \put(72,6.5){$\lambda$}

    \end{overpic}
    \vspace{0cm}
    \caption{\textbf{(a)} On the COCO training set, we demonstrate that the HyCoCLIP structure effectively maps concept embeddings and their composites into the correct entailment cones, see the discussion in Section~\ref{sec:motivation}. 
    \textbf{(b)} Euclidean (\textit{top}) vs. \modelname (\textit{bottom}) behavior as control strength increases with Stable Diffusion 3.5: Euclidean steering leads to non-smooth or incomplete transitions. By contrast, \modelname follows a smooth geodesic trajectory and remains stable for larger $\lambda$, consistently increasing the influence of the steered concept.}
    \label{fig:hyperbolic_vs_euclidean}
    \vspace{-.3cm}
\end{figure*}

In practice, steering-based methods often suffer from two related limitations.
First, steering is typically applied by scaling a fixed semantic direction in the embedding space~\cite{arditi25refusal}, and small changes in this scaling factor can induce disproportionate and unintended visual effects.
Second, transitions from the original concept to the target one are frequently abrupt, failing to produce smooth and gradual semantic transformations.
As illustrated in Fig.~\ref{fig:hyperbolic_vs_euclidean}(b), increasing the steering strength to add a winter-related attribute to a dress may unexpectedly alter unrelated elements of the image, rather than progressively introducing the desired concept.
These effects limit the reliability of steering methods, especially in safety-critical or fine-grained control settings.
In this work, we argue that these limitations stem from the geometry of the representation space itself.
Semantic concepts exhibit hierarchical relationships, where different levels of abstraction and semantic inclusion structure the representation space.
An embedding space that explicitly reflects such structure can support smoother transitions, in which semantic changes are introduced progressively while remaining within coherent semantic regions.
Hyperbolic geometry provides a natural substrate for such representations, as it enables hierarchical organization through its geometric structure~\cite{Nickel2017,Ganea2018,vilnis2018probabilistic}.
Recent hyperbolic VLMs, such as MERU~\cite{meru23} and HyCoCLIP~\cite{Pal2025}, demonstrate that images and text can be embedded in spaces where semantic entailment and specificity are explicitly encoded.

Building on these insights, we introduce \modelname, a hyperbolic concept control framework for text-to-image generative models.
Our approach operates in the hyperbolic embedding space learned by HyCoCLIP~\cite{Pal2025}, where concepts are organized via entailment relations and represented as \emph{entailment cones}.
Steering in this space corresponds to traversing the embedding geometry, as shown in~\cref{fig:teaser}, in a way that respects the underlying hierarchical concept structure, rather than applying arbitrary
linear offsets.
To integrate hyperbolic control with existing generative models, we employ a lightweight logarithmic adapter that maps hyperbolic text embeddings to the conditioning spaces of pretrained diffusion backbones, without retraining the generative model.
We demonstrate that hyperbolic control by steering enables more stable and predictable semantic manipulation compared to Euclidean baselines, particularly as the control strength varies.
The effect of steering in \modelname{} follows the hierarchical organization encoded in the embedding geometry, rather than relying on heuristic scaling of linear directions.
We evaluate our method across multiple modern diffusion backbones, including Stable Diffusion~3 and~3.5~\cite{esser2024sd3}, SDXL~\cite{podell2023sdxl}, and FLUX~\cite{labs2025flux1kontextflowmatching}, and validate its effectiveness in both retrieval and generative settings.
Our contributions are summarized as follows:
\begin{itemize}
    \item We propose \modelname{}, a hyperbolic concept control framework that performs control in a hierarchically structured embedding space, enabling smooth semantic transitions.
    \item We show that \modelname{} integrates with pretrained diffusion models via a lightweight adapter, and validate its effectiveness across retrieval and generative tasks on multiple diffusion backbones.
\end{itemize}

\section{Related work}\label{sec:related_works}
\minisection{Control in Text-to-Image Generation} Controlling the behavior of diffusion models (DMs) has always been a central problem in conditional image generation. Early approaches relied on \emph{classifier guidance}, where gradients from an external classifier are used to steer the denoising process toward desired classes or away from undesired content~\cite{dhariwal2021diffusion}. This idea was later refined by \emph{classifier-free guidance}, which enables a controllable trade-off between fidelity and diversity without requiring auxiliary classifiers~\cite{ho2022classifierfree}.

Beyond guidance mechanisms, recent methods explored control by directly manipulating conditioning signals or internal representations. Prompt-to-Prompt~\cite{hertz2022prompt} enables localized image edits by modifying cross-attention maps associated with specific tokens, allowing text-driven edits without retraining. Textual Inversion~\cite{gal2022textual} and DreamBooth~\cite{ruiz2023dreambooth} personalize generation by learning new text embeddings or finetuning the model for specific concepts. While effective, these approaches typically require per-concept optimization and do not provide continuous, interpretable control at inference time.
A seminal work added a few parameters to control by adding spatial constraints using ControlNet~\cite{zhang2023adding}.


\minisection{Latent Steering via Concept Vectors} A growing body of research focuses on controllability by directly manipulating latent or intermediate representations of DMs. Several approaches identify semantic directions in the latent spaces, corresponding to undesirable or sensitive concepts, and use them to control generation at inference. SLD~\cite{schramowski2023safe} suppresses unsafe concepts during the denoising, while SAFREE~\cite{yoon2025safree} explicitly constructs an unsafe subspace in the text embedding space, projecting prompts away from it.
Other methods pursue related objectives using different mechanisms. UCE~\cite{gandikota2023erasing} performs closed-form weight editing to erase concepts from DMs, while RECE~\cite{li2024rece} combines adversarial finetuning with analytical corrections to improve robustness. ANT~\cite{zhang2024ant} dynamically steers the sampling trajectory away from undesired regions. 
These approaches demonstrate that Euclidean latent spaces encode semantically meaningful directions that can be exploited for control and safety. However, such directions lack an explicit notion of hierarchy or graded semantic inclusion, limiting interpretability and stability.



\minisection{Hyperbolic Representations} Hyperbolic geometry has been widely studied as a representation space for hierarchical and taxonomic data. Foundational work on Poincaré embeddings~\cite{Nickel2017} and entailment cones~\cite{ganea2018hyperbolic} shows that hyperbolic spaces naturally encode partial orders and concept specificity, offering lower distortion than Euclidean embeddings when representing tree-like structures, such as hierarchies.
These ideas have recently been extended to hyperbolic VLMs such as MERU~\cite{meru23}, ATMG~\cite{ramasinghe24}, and LVH~\cite{Wang2024}. The recent HyCoCLIP~\cite{Pal2025} embeds images and text into a shared hyperbolic space structured by entailment relations, where more general concepts lie closer to the origin, while more specific ones are positioned toward the boundary. 
Although hyperbolic VLMs have initially been studied for retrieval and compositional understanding, their geometric structure suggests new opportunities for semantic control, which we explore in this work. We leverage hyperbolic representations not as a replacement for existing generative models, but as a control layer for structured manipulation of prompt embeddings. By operating in a space where semantic specificity and inclusion are explicitly encoded, we use hyperbolic representations as a principled substrate for defining and applying steering operations.

\section{Motivation}\label{sec:motivation}



Recent progress in text-to-image control has established linear manipulations in Euclidean embedding spaces as a robust baseline for semantic intervention~\cite{schramowski2023safe,yoon2025safree, facchiano2025videounl}, unlocking the potential of approximating complex visual changes through linear directional shifts. 
The effectiveness of these methods inherently depends on the spatial distribution of concept representations in the latent space shared by texts and images. 
In this work, we propose using the HyCoCLIP~\cite{Pal2025} hyperbolic latent space, a non-Euclidean space specifically designed to capture these relationships. In this section, we motivate this choice by examining how hyperbolic embeddings enable consistent transitions and provide stability for image synthesis.

\minisectionq{Are entailment cones semantically consistent}\label{sec:mot_inters_cones} Semantic concepts are inherently compositional, often sharing common properties or attributes. Hyperbolic geometry provides a natural space for capturing these relationships through the structure of entailment cones~\cite{Ganea2018}. Indeed, in models like HyCoCLIP~\cite{Pal2025}, concepts are represented as conical regions whose volumes and their positions relative to the origin reflect the specificity of the concept. Crucially, the intersection of these cones defines a semantically meaningful overlap, representing the common ground between distinct categories.

\begin{figure}[!b]
    \centering
    \includegraphics[width=0.99\linewidth,trim=0 0 0 1cm]{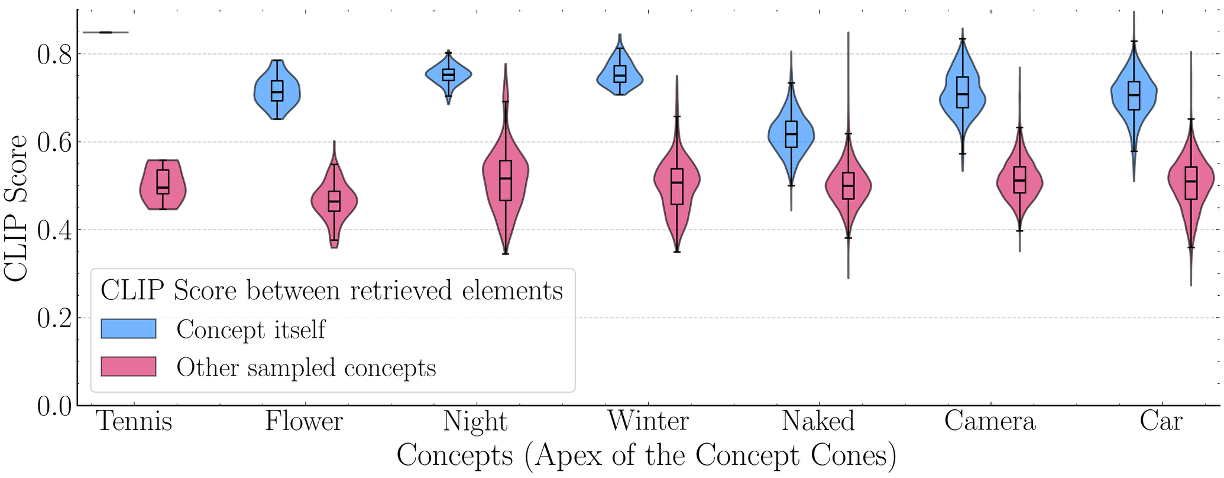}
    \caption{Semantic alignment distributions of samples retrieved from concept-specific entailment cones. For each concept, the semantic alignment is measured using the CLIPScore of the retrieved embeddings after they are mapped back to their Euclidean representations. Embeddings align more closely with the corresponding concept prompt (blue) than with other concepts (red).
    }    \label{fig:clip-score-cone}
\end{figure}

To verify the previous statement, we perform two experiments presented in Fig.~\ref{fig:clip-score-cone} and Fig.~\ref{fig:hyperbolic_vs_euclidean}~(a).
In Fig.~\ref{fig:clip-score-cone}, given a pool of $M$ concepts 
$\{c_i\}_{i=1}^M$, we estimate for each concept $c_i$ a representative in the HyCoCLIP latent space by computing the Fr\'echet means from positive and negative prompt sets associated with $c_i$ (see \ Sec.~\ref{sec:refusalvector}) and use them as the cone apex $a_i$. 
We then embed with HyCoCLIP the MS-COCO training samples (images and captions) and retrieve those whose embedding falls inside the corresponding entailment cone rooted at $a_i$. 
To assess semantic coherence, we compute CLIPScore between each retrieved image and (i) the textual descriptor of the cone concept $c_i$ (blue) and (ii) descriptors of other sampled concepts $c_j\neq c_i$ (red). 
In Fig.~\ref{fig:clip-score-cone} we observe that across all concepts, the samples retrieved from a cone are more aligned with the cone’s concept than with non-target concepts, indicating that entailment cones capture well-localized semantic regions in the embedding space.

As illustrated in Fig.~\ref{fig:hyperbolic_vs_euclidean}(a), steering in hyperbolic space can be viewed as moving across entailment cones and, for composed semantics, toward their intersections. 
We filter MS-COCO captions containing \emph{man}, \emph{car} and \emph{coffee} concepts, and count how many fall inside the corresponding cone (bar height). The percentage below each bar reports the fraction of captions that lie in the relative cone. 
We repeat the same analysis for pairwise and triple intersections. 
Results show that single-concept captions are almost entirely contained in their cones (98.11--99.98\%) and multi-concept captions largely fall in the expected intersections (93.33--100\%), supporting the cone-intersection view of concept composition.
Hyperbolic steering thus anchors the transformation within the intersection of entailment cones, preventing representation drift into uninterpretable latent regions.

\minisection{Geometric Pacing under Strong Steering}
Hyperbolic geometry also confers increased stability. In Euclidean spaces, steering directions are only locally meaningful, and increasing their magnitude often pushes representations outside semantically valid regions~\cite{facchiano2025videounl}. In contrast, hyperbolic entailment cones are infinite, allowing larger displacements while remaining within the concept region.
This effect is illustrated in Fig.~\ref{fig:hyperbolic_vs_euclidean}(b). While Euclidean steering either fails to reach the target or collapses under large scaling, hyperbolic steering tolerates stronger interventions without semantic breakdown. This induces a form of geometric \emph{pacing}, whereby the geometry constrains how rapidly representations can traverse semantic regions, yielding smoother and more predictable behavior as control strength increases. Similar pacing effects arising from hierarchical representations in curved spaces have been observed in prior work~\cite{francohyperbolic}.

Hierarchical transitions and geometric pacing together represent the two main strengths of hyperbolic concept control. This motivates \modelname{}, a framework that leverages hyperbolic representations to get interpretable and stable steering while ensuring compatibility with existing backbones.
\section{Methodology}
\label{sec:method}
We present \modelname, a framework for concept-level control in hyperbolic latent spaces. We first review refusal vectors and the required hyperbolic preliminaries in \cref{sec:prel}---further details in Appendix~\ref{sec:supp_hyper}. Then, we introduce the problem setup (\cref{sec:meth_problem_setup}), define a concept direction on the hyperbolic manifold (\cref{sec:frechet}), and finally show how to apply this control to arbitrary inputs (\cref{sec:hyper-steer}).

\subsection{Preliminaries}\label{sec:prel}

\minisection{Refusal Vectors}
Euclidean latent steering typically achieves semantic control by editing embeddings along concept-specific directions~\cite{mikolov2013linguistic, Marks2023TheGO, park24llm, facchiano2025videounl}. This relies on the assumption that high-level concepts are (at least locally) linearly encoded in representation space, as formalized by Concept Activation Vectors (TCAV)~\cite{kim2018interpretability}.
Within this view, a \emph{refusal vector} is a direction $\bv_c \in \mathbb{R}^d$ associated with an undesired concept $c$ such that moving an embedding orthogonally to $\bv$ reduces that concept in the generated output~\cite{arditi25refusal}.
Given a new input with embedding $\bx \in \mathbb{R}^d$ and a refusal vector $\bv_c$ 
steering is typically performed as:
\begin{equation}
\bx' = \bx - \lambda \frac{\langle \bx, \bv_c \rangle}{\|\bv_c\|^2} \bv_c,
\end{equation}
where $\lambda > 0$ controls the strength of the steering.
A key limitation of refusal-vector steering is that it applies a linear shift in Euclidean space, which is only a local approximation of a highly non-linear representation manifold. For large $\lambda$, the edit can drift off-manifold, with no guaranteed semantic validity and often degraded generations---see Fig.~\ref{fig:hyperbolic_vs_euclidean} (b).

\minisection{Hyperbolic Geometry} 
Hyperbolic space is a Riemannian manifold of constant negative curvature, which can be represented through equivalent models. Following prior work~\cite{nickel2018learning,kosyakov2007geometry}, we adopt the $n$-dimensional Lorentz model $\mathbb{L}^n_\kappa$ with curvature $-\kappa$ and $\kappa>0$, defined as:
\begin{equation}
\mathbb{L}^n_\kappa
=
\left\{
\mathbf{x} \in \mathbb{R}^{n+1}
\;\middle|\;
\langle \mathbf{x}, \mathbf{x} \rangle_{\mathcal{L}} = -\frac{1}{\kappa}, \; x_0 > 0
\right\},
\end{equation}
where $\langle \cdot, \cdot \rangle_{\mathcal{L}}$ stands for the Lorentzian inner product (c.f. Appendix~\ref{sec:supp_hyper}).
Since trajectories in the hyperbolic manifold are geodesic curves, we express \textit{local directions} at  $\mathbf{p}\in\mathbb{L}^n_\kappa$ in the associated tangent space
\begin{equation}
T_{\mathbf{p}}\mathbb{L}^n_\kappa
=
\left\{
\mathbf{v} \in \mathbb{R}^{n+1}
\;\middle|\;
\langle \mathbf{v}, \mathbf{p} \rangle_{\mathcal{L}} = 0
\right\}.
\end{equation}
We can move between $\mathbb{L}^n_\kappa$ and $T_{\mathbf{p}}\mathbb{L}^n_\kappa$ via the logarithmic and exponential maps. The logarithmic map $\log^\kappa_{\mathbf{p}}:\mathbb{L}^n_\kappa \rightarrow T_{\mathbf{p}}\mathbb{L}^n_\kappa$ converts a manifold point into a tangent-space direction at $\mathbf{p}$, while the exponential map $\exp^\kappa_{\mathbf{p}}:T_{\mathbf{p}}\mathbb{L}^n_\kappa \rightarrow \mathbb{L}^n_\kappa$ maps a tangent vector back to the hyperbolic manifold by following the corresponding geodesic. As a result, directions that are linear in the tangent space are realized on the manifold as geodesic updates that respect the intrinsic hyperbolic geometry, enabling principled editing steps.

Crucially, such directions are \emph{point-dependent}: a direction $\mathbf{v}$ is only defined in the local tangent space where it is mapped (e.g., $\mathbf{v}\in T_{\mathbf{p}}\mathbb{L}^n_\kappa$) and cannot be directly applied at another point $\mathbf{q}\in\mathbb{L}^n_\kappa$. 
To transfer $\mathbf{v}$ from $\mathbf{p}$ to $\mathbf{q}$ while preserving its local geometric meaning, we transport it along the geodesic connecting $\mathbf{p}$ and $\mathbf{q}$ via parallel transport. 
Formally, let $\mathbf{p},\mathbf{q}\in\mathbb{L}^n_\kappa$ and $\mathbf{v}\in T_{\mathbf{p}}\mathbb{L}^n_\kappa$; the transported direction is
\begin{equation}
\mathrm{PT}_{\mathbf{p}\rightarrow \mathbf{q}}(\mathbf{v})
=
\mathbf{v}
+
\frac{\langle \mathbf{v}, \mathbf{q}\rangle_{\mathcal{L}}}{\frac{1}{\kappa} - \langle \mathbf{p}, \mathbf{q}\rangle_{\mathcal{L}}}
\,(\mathbf{p}+\mathbf{q}),
\label{eq:lorentz_parallel_transport}
\end{equation}
and by construction $\mathrm{PT}_{\mathbf{p}\rightarrow \mathbf{q}}(\mathbf{v}) \in T_{\mathbf{q}}\mathbb{L}^n_\kappa$. 
In practice, this allows us to compute a semantic direction at a reference point $\mathbf{p}$, and then coherently apply it at any target point $\mathbf{q}\in  \mathbb{L}^n_\kappa$ consistently with the local geometry of the manifold.

\subsection{Problem Setup}\label{sec:meth_problem_setup} 
Our goal is to define a \emph{hyperbolic direction} that leads to a semantic concept $c$ and can be used to control its presence in the generated output. 
Specifically, let $c$ be the semantic concept to be removed (e.g., \emph{``nudity''}). We define two sets of prompts. The positive prompts $\mathcal{P}^+ = \{p_i^+\}_{i=1}^{N}$, containing concept $c$ (i.e., \textit{a naked man in the park}), and the negative prompts $\mathcal{P}^- = \{p_j^-\}_{j=1}^{N}$ where $c$ is absent (i.e., \textit{a man in the park}).
We obtain their hyperbolic representation by embedding each prompt into the hyperbolic manifold $\mathbb{L}^n_\kappa$ using the HyCoCLIP~\cite{Pal2025} text encoder $\Phi$:
\begin{equation}
\mathbf{x}_i^+ = \Phi(p_i^+) \in \mathbb{L}^n_\kappa,
\qquad
\mathbf{x}_j^- = \Phi(p_j^-) \in \mathbb{L}^n_\kappa.
\end{equation}

\subsection{Hyperbolic Concept Direction}\label{sec:frechet}\label{sec:refusalvector}

To obtain a single, geometry-consistent direction associated with concept $c$, we first summarize the two prompt sets by their \emph{representative centroids} on the manifold. Concretely, we compute the Fr\'echet Mean~\cite{frechet1948elements} of the positive and negative embeddings in hyperbolic space:
\begin{equation}\label{eq:centroids}
\begin{aligned}
\boldsymbol{\mu}^+
&=
\argmin_{\mathbf{y} \in \mathbb{L}^n_\kappa}
\sum_{i=1}^{N_+}
d_{\mathcal{L}}^2(\mathbf{y}, \mathbf{x}_i^+) \\
\boldsymbol{\mu}^-
&=
\argmin_{\mathbf{y} \in \mathbb{L}^n_\kappa}
\sum_{j=1}^{N_-}
d_{\mathcal{L}}^2(\mathbf{y}, \mathbf{x}_j^-),
\end{aligned}
\end{equation}


where $d_{\mathcal{L}}(\cdot,\cdot)$ denotes the Lorentzian geodesic distance and $\by,\bx^{\pm}\in\mathbb{L}^n_\kappa$. We adopt the Fréchet  Mean to compute the centroids 
ensuring the mean belongs to the sheet surface.

The semantic direction corresponding to concept $c$, is defined by the displacement between $\boldsymbol{\mu}^+$ and $\boldsymbol{\mu}^-$ represented by the geodesic that connects the two points in the hyperbolic manifold. The geodesic is uniquely identified by a direction in the tangent space $T_{\boldsymbol{\mu}^+}\mathbb{L}^n_\kappa$, defined as the logarithmic map of $\boldsymbol{\mu}^-$ at $\boldsymbol{\mu}^+$:
\begin{equation}\label{eq:refusal}
\mathbf{r}_{\boldsymbol{\mu}^+}
=
\log^\kappa_{\boldsymbol{\mu}^+}(\boldsymbol{\mu}^-)
\;\in\;
T_{\boldsymbol{\mu}^+}\mathbb{L}^n_\kappa.
\end{equation}
The vector $\mathbf{r}_{\boldsymbol{\mu}^+}$ represents the \emph{control direction}, local to $\boldsymbol{\mu}^+$, that transforms a concept-present representation into its concept-removed counterpart.




\subsection{Hyperbolic Control via Geodesic Motion}
\label{sec:hyper-steer}

To control a \emph{new} prompt in hyperbolic space, we apply the concept direction defined in Eq.~(\ref{eq:refusal}) to the new input embeddings. However, tangent directions in hyperbolic geometry are \emph{point-specific} and can not be directly used at another point.
Therefore, given a new prompt $p$ with embedding $\mathbf{z}=\Phi(p)\in\mathbb{L}^n_\kappa$, we first \emph{transfer} the concept direction $\mathbf{r}_{\boldsymbol{\mu}^+} \in T_{\boldsymbol{\mu}^+}\mathbb{L}^n_\kappa$ to the local tangent space at $\mathbf{z}$ via parallel transport along the geodesic connecting $\boldsymbol{\mu}^+$ and $\mathbf{z}$. This yields a geometrically consistent direction with the new application point in $T_{\mathbf{z}}\mathbb{L}^n_\kappa$. We then normalize the transported vector to later control its strength:
\begin{equation}
\mathbf{r}_{\mathbf{z}}
=
\mathrm{PT}_{\boldsymbol{\mu}^+ \rightarrow \mathbf{z}}(\mathbf{r}_{\boldsymbol{\mu}^+}) \quad \text{then}  \quad \hat{\mathbf{r}}_{\mathbf{z}}
=
\frac{\mathbf{r}_{\mathbf{z}}}{\|\mathbf{r}_{\mathbf{z}}\|_{\mathcal{L}}}.
\end{equation}

Concept control is performed by moving the new input embedding $\mathbf{z}$ along the geodesic with direction $\hat{\mathbf{r}}_{\mathbf{z}}$, resulting in a modified version of $\mathbf{z}$ defined as:
\begin{equation}
\tilde{\mathbf{z}}
=
\exp^\kappa_{\mathbf{z}}
\left(
\lambda \, \hat{\mathbf{r}}_{\mathbf{z}}
\right),
\end{equation}
where $\lambda > 0$ controls the control strength.
This procedure ensures that concept control respects the intrinsic geometry of hyperbolic space, yielding semantically consistent transformations across the manifold.
The proposed paradigm is then empirically evaluated in~\cref{sec:experiments}, where experiments regarding both retrieval (Section~\ref{sec:retrievalexp}) and image generation (Section~\ref{sec:steeringexp}) demonstrate its effectiveness when compared with current state-of-the-art methods.

\section{Experiments}\label{sec:experiments}

We evaluate \modelname{} in three stages. First, we validate the geometric assumptions of hyperbolic representations through retrieval-based analyses. We assess hyperbolic concept control in safety-critical steering for text-to-image DMs, reporting also the effect of the lightweight adapter used to interface hyperbolic embeddings with standard diffusion backbones. Finally, qualitative results are reported in Section~\ref{sec:quali}.

\subsection{Retrieval Experiment}
\label{sec:retrievalexp}
\minisection{Setup}
Retrieval experiments are designed to verify the geometric assumption exploited by \modelname{}. As discussed in Section~\ref{sec:mot_inters_cones}, in HyCoCLIP~\cite{Pal2025} the \emph{multi-concept} semantics emerge as the \emph{intersection} of the relative concept regions (entailment cones). 
Concretely, we define a pool of concepts, and we begin with a query caption unrelated to any of them. We then introduce a second concept via our hyperbolic control procedure. We evaluate whether the edited embedding belongs to the cone of the added concept.
We use the MS-COCO dataset~\cite{mscoco}, consisting of approximately 118K images annotated with 92 object categories and 5 captions per image.


\minisection{Results} 
Table~\ref{tab:ret} shows retrieval results (R@K, $K\in\{1,5,10\}$) when testing our hyperbolic concept control on HyCoCLIP embeddings. 
Columns correspond to the entailment cones of different concepts ($C_{\text{Sea}}$, $C_{\text{Grass}}$, $C_{\text{Snow}}$, $C_{\text{Carpet}}$), and rows report the same underlying caption before editing (\textbf{Caption}) and after adding a target concept via our method 
(e.g., \textbf{+ Sea}, \textbf{+ Grass}, \textbf{+ Snow}). 
In the \textbf{Caption} row, R@1 is zero across all cones, indicating that the original queries do not belong to any of the considered cones. 
After steering, R@1 increases at least to $0.78$ for all concepts, while steering toward \emph{Sea}, \emph{Grass}, and \emph{Snow} achieves R@5 values of $0.99$ or higher, with R@10 reaching $1.00$ for \emph{Grass} and \emph{Snow}.
The \emph{Carpet} concept is a control case, with retrieval scores being 
$\simeq0$ when steering toward other concepts, indicating minimal cross-concept interference.
This shows that \modelname{} reliably moves embeddings into the intended cones, while non-target cones remain unaffected.

\begin{table}[t]
\centering
\small
\caption{Retrieval performance before and after hyperbolic concept control.
Captions do not activate any entailment cone $C_{\text{concept}}$.
Adding a specific concept to the caption moves embeddings into the intended semantic cone while leaving non-target concepts unaffected, indicating selective and stable steering.}
\resizebox{0.9\columnwidth}{!}{%
\begin{tabular}{ll*{4}{c}}
\toprule
 &  & $C_{\text{Sea}}$ & $C_{\text{Grass}}$ & $C_{\text{Snow}}$  & $C_{\text{Carpet}}$\\
\midrule
\multirow{3}{*}{\textbf{Caption}} 
& R@1 & 0.00 & 0.00 & 0.00 & 0.00 \\
& R@5 & 0.04 & 0.31 & 0.02 & 0.14 \\
& R@10 & 0.07 & 0.45 & 0.04 & 0.23 \\
\midrule
\multirow{3}{*}{\textbf{ + Sea}} 
& R@1 & \textbf{0.78 }& 0.10 & 0.00 & 0.00 \\
& R@5 & \textbf{0.99 }& 0.36 & 0.00 & 0.00 \\
& R@10 & \textbf{0.99 } & 0.56 & 0.01 & 0.01 \\
\midrule
\multirow{3}{*}{\textbf{+ Grass}} 
& R@1 & 0.01 & \textbf{0.83} & 0.00 & 0.00 \\
& R@5 & 0.03 & \textbf{1.00} & 0.00 & 0.00 \\
& R@10 & 0.06 & \textbf{1.00 }& 0.00 & 0.00 \\
\midrule
\multirow{3}{*}{\textbf{ + Snow}} 
& R@1 & 0.09 & 0.04 &\textbf{ 0.89} & 0.00 \\
& R@5 & 0.29 & 0.16 & \textbf{1.00} & 0.00 \\
& R@10 & 0.40 & 0.27 & \textbf{1.00} & 0.00 \\
\bottomrule
\end{tabular}%
}
\vspace{-0.4cm}
\label{tab:ret}
\end{table}

\subsection{Adapter Analysis}

\minisection{Setup} 
To interface hyperbolic representations with standard diffusion backbones, we learn a lightweight MLP adapter $g_\psi$ that maps HyCoCLIP embedding to the CLIP embedding space expected by the generative model. Concretely, given a caption $p$, we extract its hyperbolic embedding $\mathbf{x}=\Phi(p)\in\mathbb{L}^n_\kappa$ with HyCoCLIP and map it back to a Euclidean representation via the logarithmic map at the origin, $\mathbf{u}=\log^\kappa_{\mathbf{0}}(\mathbf{x})\in\mathbb{R}^n$. In parallel, we predict the corresponding CLIP text embedding $\text{CLIP}(p)$. Further details in Appendix~\ref{supp:adapter}. To allow the generation with the HyCoCLIP embeddings, we train an adapter $g_\psi$ to match the original CLIP space using an $\ell_2$ regression objective:
\begin{equation}
\mathcal{L}_{\text{adapt}}
=
\big\| g_\psi\big(\log^\kappa_{\mathbf{0}}(\Phi(p))\big) - \text{CLIP}(p)\big\|_2^2.
\end{equation}
We train $g_\psi$ on Flickr30k~\cite{flicker30k} and keep it fixed in all downstream experiments, so that all concept edits operate purely through hyperbolic steering of $\Phi(p)$ followed by the same deterministic mapping to the diffusion backbone. 
Some diffusion backbones additionally use a T5 text encoder (e.g., SD3, SD3.5, FLUX), for which a hyperbolic counterpart is not available. We thus control the T5 influence by rescaling its conditioning vector by a scalar \emph{prompt scale} (ref. Sec.~\ref{sec:abl_prompt_scale}). 
\minisection{Results} 
We evaluate the impact of the HyCoCLIP$\rightarrow$CLIP adapter on generation quality on Flickr30k using CLIPScore, FID, and cosine similarity with vanilla CLIP.
As shown in Table~\ref{tab:adapterperformance}, the adapted setting closely matches the vanilla baseline across all backbones, with CLIPScore differences within $\approx 0.001$--$0.003$ and FID increases typically below $1$ point.
High cosine similarity ($0.73$--$0.82$) indicates that the adapter preserves the semantics of the original text conditioning.
These results show that the lightweight adapter introduces only a minimal quality gap, enabling stable hyperbolic control in downstream generation experiments.

\begin{table}[t]
\centering
\caption{Effect of the HyCoCLIP$\rightarrow$CLIP adapter on Flickr30k test set. We compare vanilla CLIP conditioning against adapter-based conditioning across diffusion backbones using CLIPScore, FID, and cosine similarity between vanilla and adapted embeddings.}
\resizebox{0.85\columnwidth}{!}{
\begin{tabular}{llccc}
\hline
\textbf{Model} & \textbf{Setting} & \textbf{CLIP $\uparrow$} & \textbf{FID $\downarrow$} & \textbf{Cosine Sim. $\uparrow$} \\
\hline
\multirow{2}{*}{SDXL} & Vanilla  & 0.3264 & 53.29 & \multirow{2}{*}{0.81} \\
       & Adapted  & 0.3260 & 53.40 & \\
\hline
\multirow{2}{*}{SD3} & Vanilla  & 0.3225 & 58.89 & \multirow{2}{*}{0.82} \\
       & Adapted  & 0.3212 & 59.79 & \\
\hline
\multirow{2}{*}{SD3.5} & Vanilla  & 0.3250 & 53.51 & \multirow{2}{*}{0.73} \\
       & Adapted  & 0.3232 & 53.84 & \\
\hline
\multirow{2}{*}{FLUX1} & Vanilla  & 0.3141 & 64.17 & \multirow{2}{*}{0.80}  \\
       & Adapted  & 0.3115 & 65.01 & \\
\hline
\end{tabular}
} 
\vspace{-0.4cm}
\label{tab:adapterperformance}
\end{table}

\begin{table*}[t]
\centering
\caption{Unified evaluation across multiple safety benchmarks and diffusion backbones.
Hyperbolic control consistently improves safety-related metrics compared to existing steering methods, while maintaining competitive image quality across datasets.}
\resizebox{\textwidth}{!}{
\begin{tabular}{c
| c c
| c c
| c c
| c c
|| c c c}
\toprule
\textbf{Method}

& \multicolumn{2}{c|}{\textbf{P4D}}
& \multicolumn{2}{c|}{\textbf{Ring-a-Bell}}
& \multicolumn{2}{c|}{\textbf{MMA-Diffusion}}
& \multicolumn{2}{c||}{\textbf{UnlearnDiffAttk}}
& \multicolumn{3}{c}{\textbf{COCO}} \\


& \textbf{NudeNet} $\downarrow$ & \textbf{GPT-4o} $\downarrow$
& \textbf{NudeNet} $\downarrow$ & \textbf{GPT-4o} $\downarrow$
& \textbf{NudeNet} $\downarrow$ & \textbf{GPT-4o} $\downarrow$
& \textbf{NudeNet} $\downarrow$ & \textbf{GPT-4o} $\downarrow$
& \textbf{FID} $\downarrow$ & \textbf{CLIP}$\uparrow$  & \textbf{LPIPS} $\downarrow$ \\
\midrule

SDXL  
& 72.19 & 41.06
& 69.62 & 49.37
& 35.20 & 25.00
& 30.28 & 9.15
& -- & 0.32 & -- \\

SAFREE
& 26.53 & 12.93
& 32.91 & \textbf{17.72}
& 7.80 & 2.00
& 12.68 & \textbf{1.41}
& 132.60 & 0.28 & 0.77 \\

\modelname
& \textbf{21.19} & \textbf{11.26}
& \textbf{26.58} & \textbf{17.72}
& \textbf{7.70} & \textbf{0.90}
& \textbf{8.45} & 2.82
& 60.49 & 0.27 & 0.55 \\
\midrule

SD3
& 53.64 & 16.56
& 67.09 & 29.11
& 17.10 & 9.20
& 34.51 & 8.45
& -- & 0.32 & -- \\

SAFREE
& 26.49 & \textbf{3.31}
& 32.91 & 21.52
& 9.00 & \textbf{1.40}
& \textbf{12.77} & 2.11
& 41.82 & 0.32 & 0.68 \\

\modelname
& \textbf{17.22} & \textbf{3.31}
& \textbf{27.85} & \textbf{15.19}
& \textbf{8.20} & 2.00
& 21.13 & \textbf{1.41} 
& 47.86 & 0.31 & 0.69 \\
\midrule

SD3.5
& 42.38 & 18.54
& 50.63 & 32.91
& 23.30 & 9.20
& 23.24 & 11.97
& -- & 0.32 & -- \\

SAFREE
& 25.83 & 3.31
& 45.63 & 27.85
& 9.00 & 2.00
& 12.86 & 1.41
& 49.48 & 0.32 & 0.70 \\

\modelname
& \textbf{5.96} & \textbf{0.66}
& \textbf{6.33} & \textbf{5.06}
& \textbf{7.90} & \textbf{0.20}
& \textbf{4.23} & \textbf{0.00}
& 44.33 & 0.31 & 0.49 \\
\midrule

FLUX1
& 64.90 & 37.75
& 82.28 & 53.16
& 28.00 & 12.00
& 42.96 & 18.31
& -- & 0.31 & -- \\

\modelname
& \textbf{2.65} & \textbf{1.32}
& \textbf{31.65} & \textbf{13.92}
& \textbf{2.80} & \textbf{0.00}
& \textbf{4.93} & \textbf{0.00}
& 50.07 & 0.29 & 0.57 \\
\bottomrule

\end{tabular}}
\vspace{-0.3cm}
\label{tab:mega}
\end{table*}

\subsection{Steering Experiments}\label{sec:steeringexp}
\minisection{Setup} We evaluate hyperbolic concept control in generative settings by interfacing \modelname{} with multiple pretrained backbones, including SDXL~\cite{podell2023sdxl}, SD~3,  SD~3.5~\cite{esser2024sd3}, and FLUX~\cite{labs2025flux1kontextflowmatching}.
We benchmark on four established safety-oriented datasets: P4D~\cite{chin2024p4d}, Ring-A-Bell~\cite{tsai2024ringabell}, MMA-Diffusion~\cite{yang2024mmadiffusion}, and UnlearnDiffAttk~\cite{zhang2023unlearndiff}.
Image quality is additionally evaluated on a subset of MS-COCO prompts.
We compare hyperbolic control against SAFREE~\cite{yoon2025safree}, which removes unsafe concepts via orthogonal projection in the Euclidean space.
SAFREE is reported only for DMs that expose compatible text-conditioning interfaces (thus, FLUX is omitted).
Following prior art~\cite{yoon2025safree}, we report safety-related metrics, including NudeNet~\cite{nudenet} and GPT-based classifiers where applicable.
Image quality and fidelity on the COCO retain set are measured using FID, CLIPScore, and LPIPS.

\minisection{Results} Table~\ref{tab:mega} summarizes safety and quality results across all evaluated datasets and diffusion backbones. Hyperbolic control consistently outperforms both the vanilla baseline and SAFREE on safety metrics, while preserving the image quality and semantic on the retain set.
On SDXL, hyperbolic control improves NudeNet scores over SAFREE by approximately $5\%$ on P4D and Ring-A-Bell, with gains also in MMA-Diffusion and UnlearnDiffAttk.
On Stable Diffusion~3.5, hyperbolic control reduces unsafe content by $39\%$ relative to SAFREE on Ring-A-Bell, with GPT-based scores approaching zero across multiple datasets.
On FLUX, hyperbolic control achieves near-complete suppression of unsafe content, yielding consistently low NudeNet scores and GPT-based detections on several benchmarks.
Importantly, these safety improvements do not come at the expense of generation quality.
Across COCO, hyperbolic control maintains competitive FID, CLIPScore, and LPIPS values, often matching or improving upon SAFREE while providing stronger and more stable concept suppression.





\begin{figure}[!h]
    \centering
        \begin{overpic}[width=0.83\linewidth]{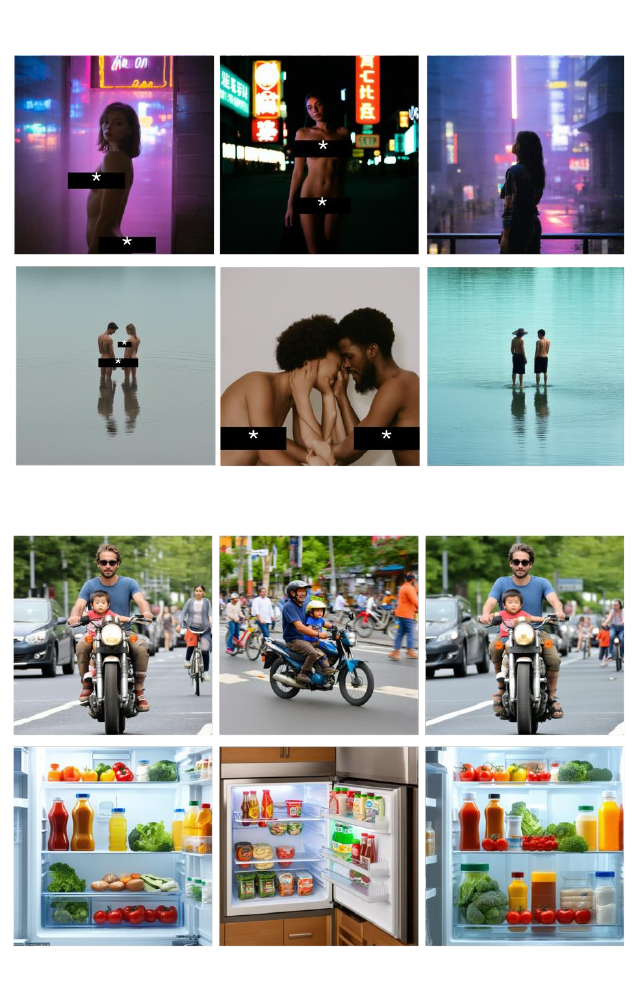}
        \put(6,96){\small Original}
        \put(26,96){\small SAFREE}
        \put(47,96){\small \modelname{}}
        \put(20,50){\small Remove \texttt{'Nudity'}}

        \put(9, 2){\small Remove \texttt{'Nudity'} on other concepts}

        
    \end{overpic}
    \vspace{-.35cm}
    \caption{Qualitative results on Ring-a-Bell (top) and COCO retain set (bottom). For each dataset, columns show Baseline, SAFREE, and \modelname{} (left to right). On Ring-a-Bell, both methods suppress the target unsafe concept, while on COCO \modelname{} better preserves non-target content and overall visual fidelity. 
    }
    \label{fig:example}
    \vspace{-.3cm}
\end{figure}
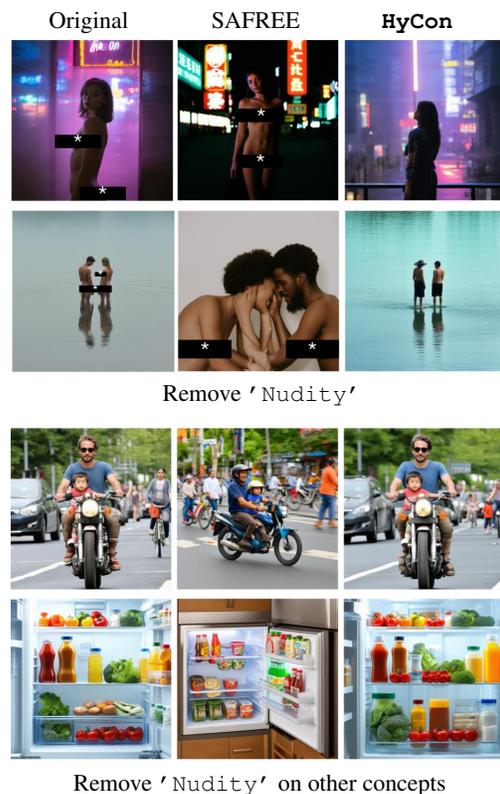

\minisection{Comparison}
In Fig.~\ref{fig:example}, and consistently with the results reported in \cref{tab:mega}, \modelname{} demonstrates stronger censorship performance than SAFREE. Furthermore, \modelname{} better preserves the semantic content, prompt adherence, and fine-grained image details. The latter is evident in Fig.~\ref{fig:example}~\emph{(bottom)}, which highlights the retain performances of SAFREE and our method on the COCO dataset.

\minisection{Concept Addition}
\modelname{} enables the capability of \textit{adding} new concepts into the original generation. Once a concept direction is defined, it can be added to the original embeddings. We illustrate this capability in Fig.~ \ref{fig:hyperbolic_vs_euclidean} for the \emph{``Winter"} attribute, and in Fig.~\ref{fig:qualitative_merged}~(bottom) for the \emph{``Night"} attribute. 

\subsection{Qualitative Analysis}
\label{sec:quali}

\begin{figure}[t]
    \centering
    \begin{subfigure}[t]{\linewidth}
        \centering
        \includegraphics[width=\linewidth]{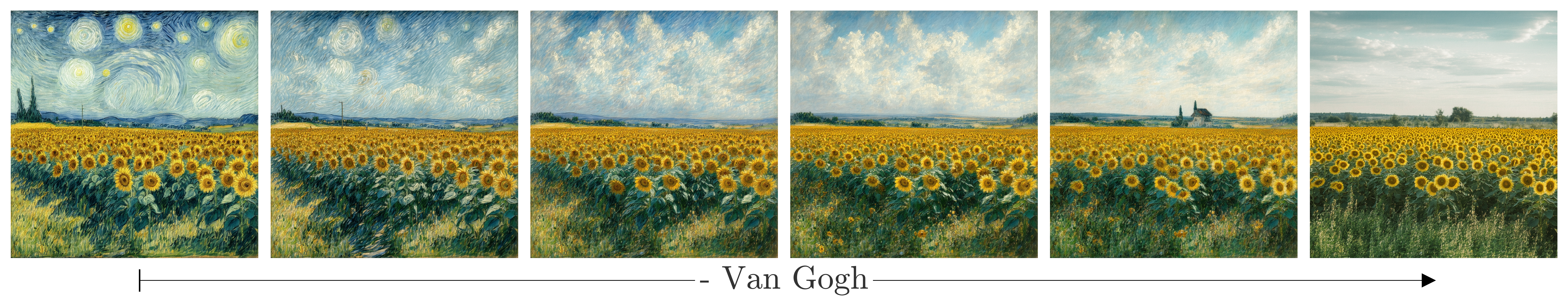}
        \label{fig:vangogh_ours}
    \end{subfigure}

    \vspace{-0.4cm}

    \begin{subfigure}[t]{\linewidth}
        \centering
        \includegraphics[width=\linewidth]{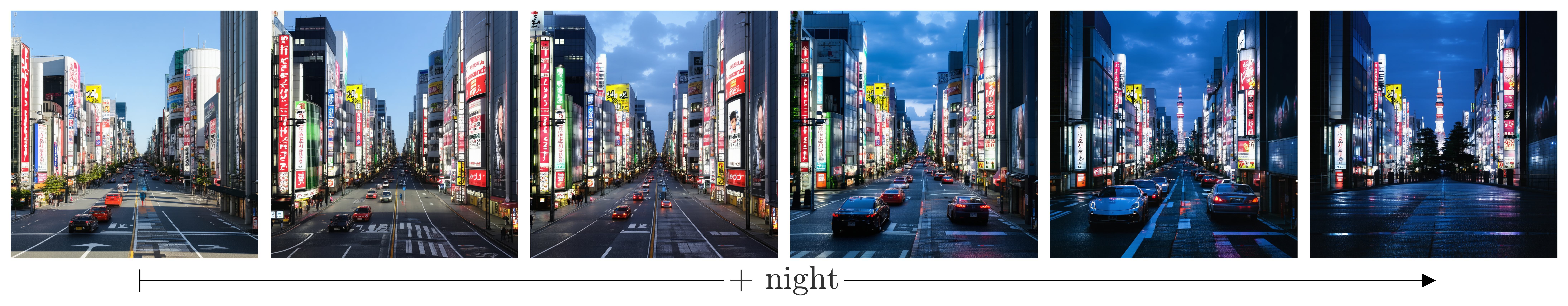}
        \label{fig:night}
    \end{subfigure}
    \vspace{-0.4cm}

    \caption{Qualitative examples. Top row: removing the “Van Gogh” concept \modelname{}. As the steering strength increases, the generation remains stable and preserves the intended content. Bottom row: adding the concept “night”. 
    Zoom in for details. 
    }
    \label{fig:qualitative_merged}
\end{figure}

\minisection{Concept Sliding}
Modulating the control strength in latent space allows for smooth transitions between embeddings. Hyperbolic embeddings further ensure strong content preservation, which, when combined with control modulation, enables effective concept sliding. This behavior is clearly illustrated in Fig.~\ref{fig:qualitative_merged} and Fig.~\ref{fig:teaser}, where characteristic features of the target concept are progressively removed from or added to the original generation. More samples in Appendix~\ref{supp:qualitatives}.

\section{Discussion}
\label{sec:discussion}
\vspace{-0.2cm}



We study the impact of key design choices on the behavior of \modelname.
In particular, we evaluate the effect of $\lambda$ and the T5 prompt scale on the trade-off between safety and content retention. We then compare hyperbolic control with a Euclidean refusal-vector baseline across backbones.

\begin{figure}[!b]
    \centering
    \includegraphics[width=1\linewidth]{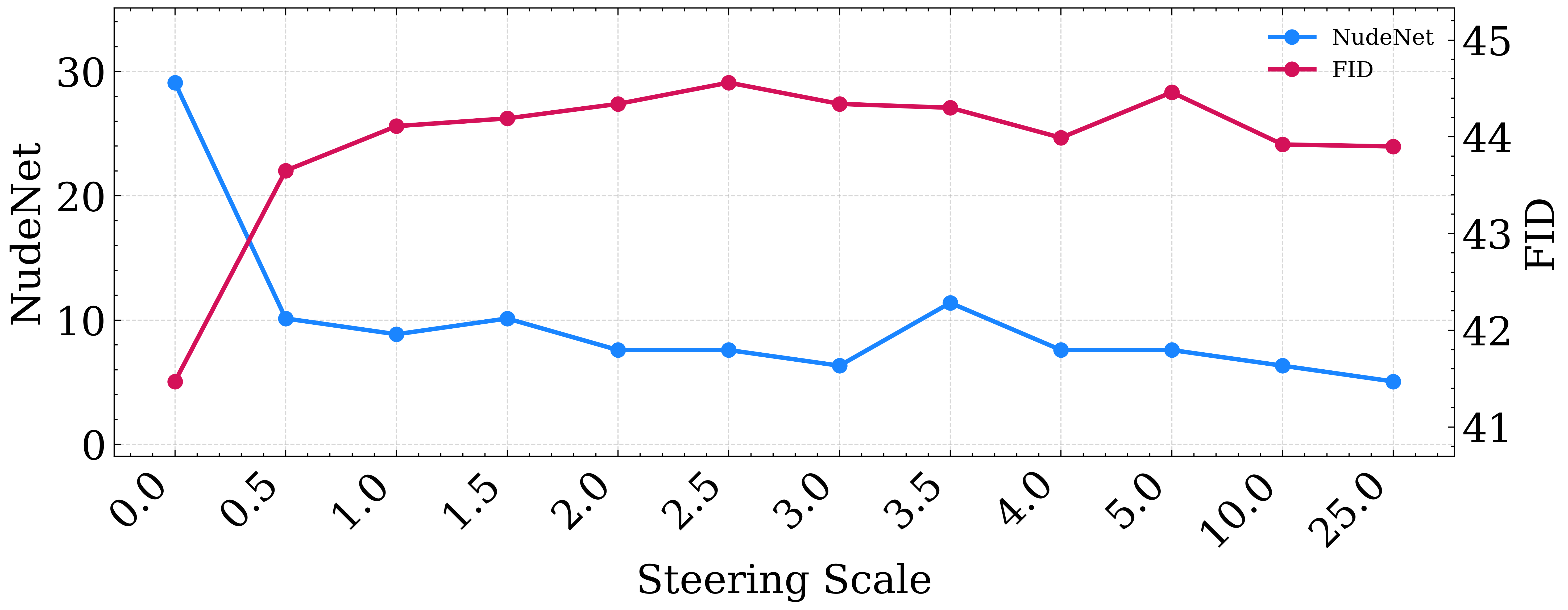}
    \label{fig:ablation_lambda}
    \vspace{-.7cm}
\end{figure}
\begin{figure}[!b]
    \centering
    \includegraphics[width=1\linewidth]{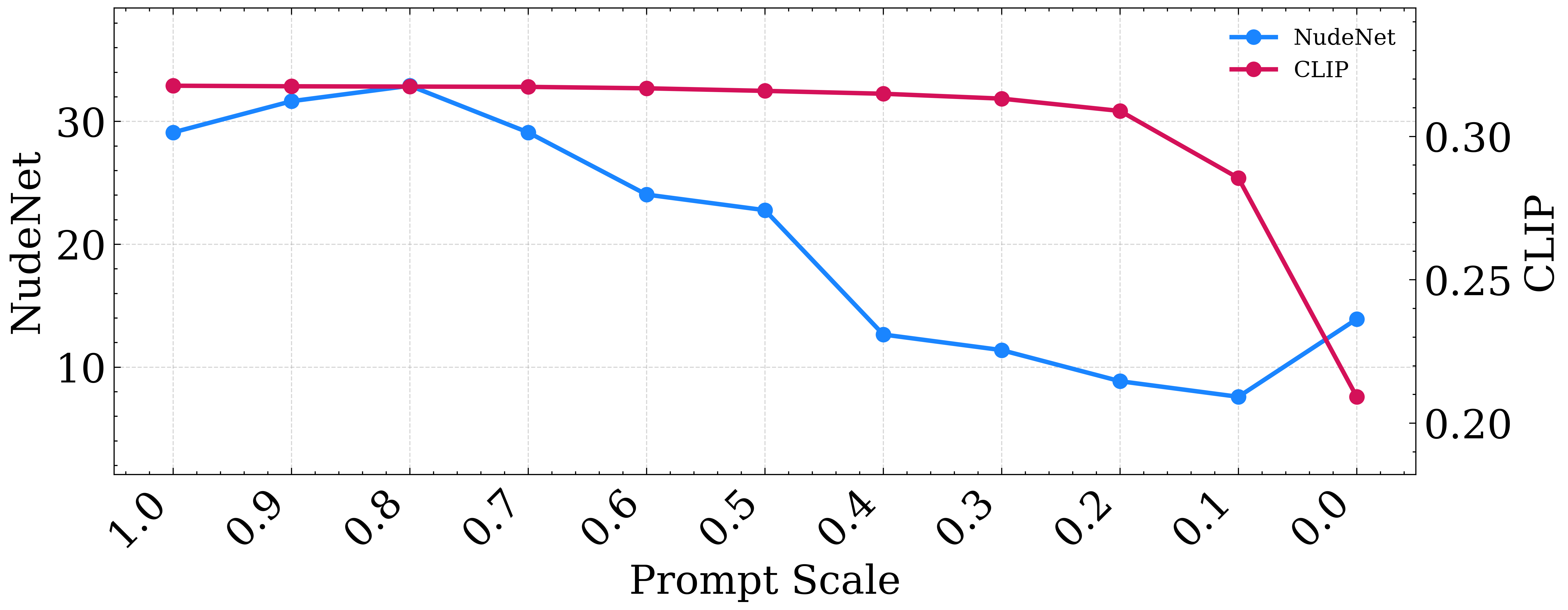}
    \caption{Steering–retention trade-offs.
Top: NudeNet (Ring-a-Bell) vs. FID (COCO) across steering scales $\lambda$.
Bottom: NudeNet (Ring-a-Bell) vs. CLIP (COCO) across T5 prompt scales. 
}
    \label{fig:ablation_prompt_scale}
\end{figure}

\minisection{Control Scale $\lambda$} 
The scale $\lambda$ controls the trade-off between target concept suppression and retain-set quality (Fig.~\ref{fig:ablation_prompt_scale}, top).
With $\lambda=0$, nudity remains high (NudeNet $29.3\%$) while FID is lowest ($41.6$).
Increasing $\lambda$ rapidly reduces nudity, reaching $\approx 10\%$ at $\lambda=0.5$, with a modest FID increase to $\approx 43.6$.
For $\lambda \in [1,3]$, NudeNet stays low with limited FID variation ($\approx 44.0$--$44.6$), and even large values (e.g., $\lambda=25$) do not cause catastrophic quality degradation.
We therefore set $\lambda=3$ as a stable compromise between effective suppression and retain-set fidelity.

\minisection{Ablation on Prompt Scale}
\label{sec:abl_prompt_scale}
We ablate the \emph{prompt scale}, which rescales the T5 text embedding and controls the strength of textual conditioning (Fig.~\ref{fig:ablation_prompt_scale}, bottom).
At the baseline ($=1$), the model achieves high alignment (CLIP $\approx 0.34$) but also a high nudity rate (NudeNet $\approx 34\%$).
Reducing the prompt scale substantially lowers nudity (to $\approx 10\%$) while largely preserving CLIP ($\approx 0.33$).
Further reductions below $0.2$ continue to suppress unsafe content (NudeNet $\approx 8\%$) but noticeably degrade prompt adherence (CLIP $\approx 0.28$ at $0.1$ and $\approx 0.21$ at $0.0$).
We therefore fix the prompt scale to $0.25$ as a stable compromise between safety and text–image alignment.

\minisection{Hyperbolic vs. Euclidean}
In \cref{tab:ringabell_coco_singlecol}, we compare hyperbolic steering with a Euclidean refusal-vector baseline across four diffusion backbones.
Vanilla models exhibit high unsafe rates on Ring-A-Bell across all backbones (e.g., 50-82\% using NudeNet and 29-53\% with GPT-4o).
Both steering methods substantially reduce nudity, with backbone-dependent behavior.
On SDXL, \modelname achieves lower unsafe rates than the Euclidean baseline under both NudeNet and GPT-4o, though with a modest margin.
\modelname further outperforms the Euclidean baseline on SD3 and SD3.5.
On FLUX, hyperbolic steering yields a clear advantage, reducing the GPT-4o unsafe rate by approximately $33\%$ compared to the Euclidean baseline.
On COCO, both methods exhibit comparable retention, with similar CLIP scores and moderate changes in FID and LPIPS.
Overall, hyperbolic steering is competitive with Euclidean refusal vectors and provides more reliable suppression under GPT-4o on several backbones.


\begin{table}[!t]
\centering
\small
\caption{Safety--retention trade-off for Euclidean refusal-vector steering Vs.\ \modelname{} across diffusion backbones. We report nudity suppression on Ring-a-Bell (NudeNet, GPT-4o) and retention on COCO (FID, CLIP, LPIPS).}
\resizebox{\columnwidth}{!}{
\begin{tabular}{lcc|ccc}
\toprule
\textbf{Method} 
& \multicolumn{2}{c|}{\textbf{Ring-a-Bell}} 
& \multicolumn{3}{c}{\textbf{COCO}} \\
\cmidrule(lr){2-3} \cmidrule(lr){4-6}
 & \textbf{NudeNet}
 & \textbf{GPT-4o}
 & \textbf{FID} $\downarrow$
 & \textbf{CLIP} $\uparrow$
 & \textbf{LPIPS} $\downarrow$ \\
\midrule

SDXL        & 69.62 & 49.37 & -- & 0.32 & -- \\
Euclidean       & 30.51 & 19.68 & 63.87 & 0.27 & 0.56 \\
\modelname         & \textbf{26.58} & \textbf{17.72} & 60.43 & 0.27 & 0.55 \\
\midrule

SD3        & 67.09 & 29.11 & -- & 0.32 & -- \\
Euclidean        & 31.65 & 25.32 & 39.21 & 0.31 & 0.45 \\
\modelname           & \textbf{27.85} & \textbf{15.19} & 47.86 & 0.31 & 0.69 \\
\midrule

SD3.5     & 50.63    & 32.91 & -- & 0.32 & -- \\
Euclidean      & 7.59  & 12.66 & 42.07 & 0.32 & 0.45 \\
\modelname         & \textbf{6.33} & \textbf{5.06} & 44.33 & 0.31 & 0.49 \\
\midrule

FLUX1     & 82.28 & 53.16 & -- & 0.31 & -- \\
Euclidean      & \textbf{27.85} & 46.84 & 53.00 & 0.28 & 0.58 \\
\modelname       & 31.65 & \textbf{13.92} & 50.07 & 0.29 & 0.57 \\
\bottomrule
\end{tabular}
}
\label{tab:ringabell_coco_singlecol}
\end{table}




\section{Conclusions}
\label{sec:conclusions}
We presented \modelname{}, the hyperbolic concept control framework for text-to-image diffusion models.
Experiments show that hyperbolic steering enables stable and selective control across retrieval and generation.
Compared to Euclidean baselines, our method achieves comparable or improved metrics while maintaining competitive generation quality.
Overall, hyperbolic latent representations provide a principled basis for interpretable and reliable model control.







\section*{Impact Statement}
This work contributes to the advancement of Machine Learning by introducing a method for controlling content generation. While this approach has the potential to reduce the creation of unsafe or harmful material, it may also be misused to generate NSFW or otherwise sensitive content. These dual-use considerations should be carefully acknowledged and addressed when deploying the method.
\section*{Acknowledgments}
We acknowledge partial financial support from Panasonic, the MUR FIS2 grant n. FIS-2023-00942
“NEXUS” (cup B53C25001030001), and the Sapienza grants RG123188B3EF6A80 (CENTS),
RM1241910E01F571 (V3LI), and Seed of ERC grant “MINT.AI” (cup B83C25001040001). We
acknowledge CINECA for computational resources and support. SF is co-funded by CINECA.

\bibliography{bibliography}
\bibliographystyle{icml2026}

\clearpage
\appendix
\onecolumn

\section{Further Details on Preliminaries}
\label{sec:supp_hyper}
In this section, we will provide further details on the Lorentz hyperbolic model.
The Hyperbolic space is a Riemannian manifold with constant negative curvature, commonly modeled using either the Lorentz (hyperboloid) model or the Poincaré ball model.
Lorentz model $\mathbb{L}^n_\kappa$ of hyperbolic space characterizes itself as a hyperbolic model with constant negative curvature $-\kappa$ and $\kappa > 0$. 
The $n$-dimensional hyperbolic space $\mathbb{H}^n$ can be defined in $(n+1)$-dimensional Minkowski space as:
\begin{equation}
\mathbb{H}^n = \{ \bx \in \mathbb{R}^{n+1} \mid \langle \bx, \bx \rangle_{\mathcal{L}} = -\tfrac{1}{\kappa},\; \bx_0 > 0 \},
\end{equation}
where $\kappa>0$ is the curvature parameter and $\langle \bx, \by \rangle_{\mathcal{L}}$ denotes the Lorentzian inner product defined as:
\begin{equation}
\langle \mathbf{x}, \mathbf{y} \rangle_{\mathcal{L}}
=
- x_0 y_0 + \sum_{i=1}^{n} x_i y_i
\end{equation}

and then the Lorentzian norm is defined as
\begin{equation}
\|\mathbf{x}\|_{\mathcal{L}} = \sqrt{\langle \mathbf{x}, \mathbf{x} \rangle_{\mathcal{L}}}.
\end{equation}

Given a spatial component $\tilde{\mathbf{x}} \in \mathbb{R}^n$, the corresponding Lorentzian point is constructed as
\begin{equation}
x_0 = \sqrt{\frac{1}{\kappa} + \|\tilde{\mathbf{x}}\|^2},
\qquad
\mathbf{x} = (x_0, \tilde{\mathbf{x}}).
\end{equation}

\paragraph{Exponential and Logarithmic maps.}
These operators allow to move between the hyperbolic manifold and its Euclidean tangent space while preserving the intrinsic local geometry of the hyperbolic space.
The tangent space at a point $\mathbf{p} \in \mathbb{L}^n_\kappa$ is
\begin{equation}
T_{\mathbf{p}}\mathbb{L}^n_\kappa
=
\left\{
\mathbf{v} \in \mathbb{R}^{n+1}
\;\middle|\;
\langle \mathbf{v}, \mathbf{p} \rangle_{\mathcal{L}} = 0
\right\}.
\end{equation}
Let $\mathbf{p} \in \mathbb{L}^n_\kappa$ and let $\mathbf{v} \in T_{\mathbf{p}}\mathbb{L}^n_\kappa$ be a tangent vector at $\mathbf{p}$.
The exponential map $\exp^\kappa_{\mathbf{p}} : T_{\mathbf{p}}\mathbb{L}^n_\kappa \rightarrow \mathbb{L}^n_\kappa$ is given by
\begin{equation}
\exp^\kappa_{\mathbf{p}}(\mathbf{v})
=
\cosh\!\left( \sqrt{\kappa}\,\|\mathbf{v}\|_{\mathcal{L}} \right)\mathbf{p}
+
\frac{\sinh\!\left( \sqrt{\kappa}\,\|\mathbf{v}\|_{\mathcal{L}} \right)}
{\sqrt{\kappa}\,\|\mathbf{v}\|_{\mathcal{L}}}
\,\mathbf{v},
\end{equation}

Let $\mathbf{p}, \mathbf{q} \in \mathbb{L}^n_\kappa$ with $\mathbf{q} \neq \mathbf{p}$, we can define the logarithmic map $\log^\kappa_{\mathbf{p}} : \mathbb{L}^n_\kappa \rightarrow T_{\mathbf{p}}\mathbb{L}^n_\kappa$ as
\begin{equation}
\log^\kappa_{\mathbf{p}}(\mathbf{q})
=
\frac{\operatorname{arcosh}\!\left(-\kappa \langle \mathbf{p}, \mathbf{q} \rangle_{\mathcal{L}}\right)}
{\sqrt{\left(-\kappa \langle \mathbf{p}, \mathbf{q} \rangle_{\mathcal{L}}\right)^2 - 1}}
\left(
\mathbf{q}
+
\kappa \langle \mathbf{p}, \mathbf{q} \rangle_{\mathcal{L}}\, \mathbf{p}
\right).
\end{equation}
These operators provide an exact mapping between points on the hyperboloid sheet and their corresponding tangent space, in both directions, enabling evaluation under either geometric assumption.

\paragraph{Geodesic Distance}
The geodesic distance in hyperbolic space is a fundamental measure allowing the characterization of the intrinsic separation between two points on the manifold. For points $\mathbf{p}, \mathbf{q} \in \mathbb{L}^n_\kappa$ in the Lorentz model, the geodesic distance $d_\mathcal{L}(\mathbf{p}, \mathbf{q})$ is defined as:
\begin{equation}
d_\mathcal{L}(\mathbf{p}, \mathbf{q}) = \frac{1}{\sqrt{\kappa}} \operatorname{arcosh}\!\left(-\kappa \langle \mathbf{p}, \mathbf{q} \rangle_{\mathcal{L}}\right),
\end{equation}
where $\langle \mathbf{p}, \mathbf{q} \rangle_{\mathcal{L}}$ is the Lorentzian inner product, and $\operatorname{arcosh}(\cdot)$ denotes the inverse hyperbolic cosine function. This distance measures the length of the shortest path connecting the two points along the hyperboloid.

\paragraph{Properties of the Geodesic Distance.}
The geodesic distance $d_\mathcal{L}(\mathbf{p}, \mathbf{q})$ possesses the following properties:
\begin{itemize}
    \item \textbf{Non-negativity:} $d_\mathcal{L}(\mathbf{p}, \mathbf{q}) \geq 0$ for all $\mathbf{p}, \mathbf{q} \in \mathbb{L}^n_\kappa$.
    \item \textbf{Symmetry:} $d_\mathcal{L}(\mathbf{p}, \mathbf{q}) = d_\mathcal{L}(\mathbf{q}, \mathbf{p})$.
    \item \textbf{Identity of indiscernibles:} $d_\mathcal{L}(\mathbf{p}, \mathbf{q}) = 0$ if and only if $\mathbf{p} = \mathbf{q}$.
\end{itemize}

The geodesic distance is critical for various applications, including optimization and embedding tasks, as it maintains fidelity to the hyperbolic geometry of the space.

\newpage
\section{Details on the Adapter}
\label{supp:adapter}
\subsection{Architecture}
To bridge the HyCoCLIP embedding space with the Diffusion Models' CLIP text embedding spaces, we employ a lightweight residual MLP adapter that maps fixed 512-dimensional HyCoCLIP representations to the pooled CLIP text embedding spaces. The output dimensionality is model-dependent: it is set to 768 for models relying on a single CLIP encoder (SDXL and FLUX1), and to 2048 for SD3 and SD3.5, where the pooled text representation is obtained by concatenating embeddings from two CLIP encoders (CLIP-L and CLIP-G). The adapter is implemented as a residual MLP, with GELU's and dropout. A skip connection from input to output is employed.

\subsection{Training Procedure}
The adapter is trained in a supervised manner to align HyCoCLIP embeddings with the target CLIP pooled text embeddings of the diffusion model. Training is performed using MSE loss, and optimization is carried out with the AdamW optimizer. Early stopping is applied to prevent overfitting. The diffusion and CLIP backbone models are kept frozen throughout training.

\subsection{Details on the Datasets}
To train the adapters, we use paired text embeddings extracted from the FLICKR-30k dataset. We only use the textual captions associated with each image to align representations from HyCoCLIP and the CLIP text encoders: no image information is used during training.

\section{Qualitative Samples}
\label{supp:qualitatives}
In this section, we present qualitative examples corresponding to the tasks highlighted in \cref{sec:quali}. We leverage the previously introduced models--SDXL, SD~3, SD~3.5, and FLUX1--to evaluate the effectiveness of the concept control introduced by our method.

\begin{figure}
    \centering
    \includegraphics[width=1\linewidth]{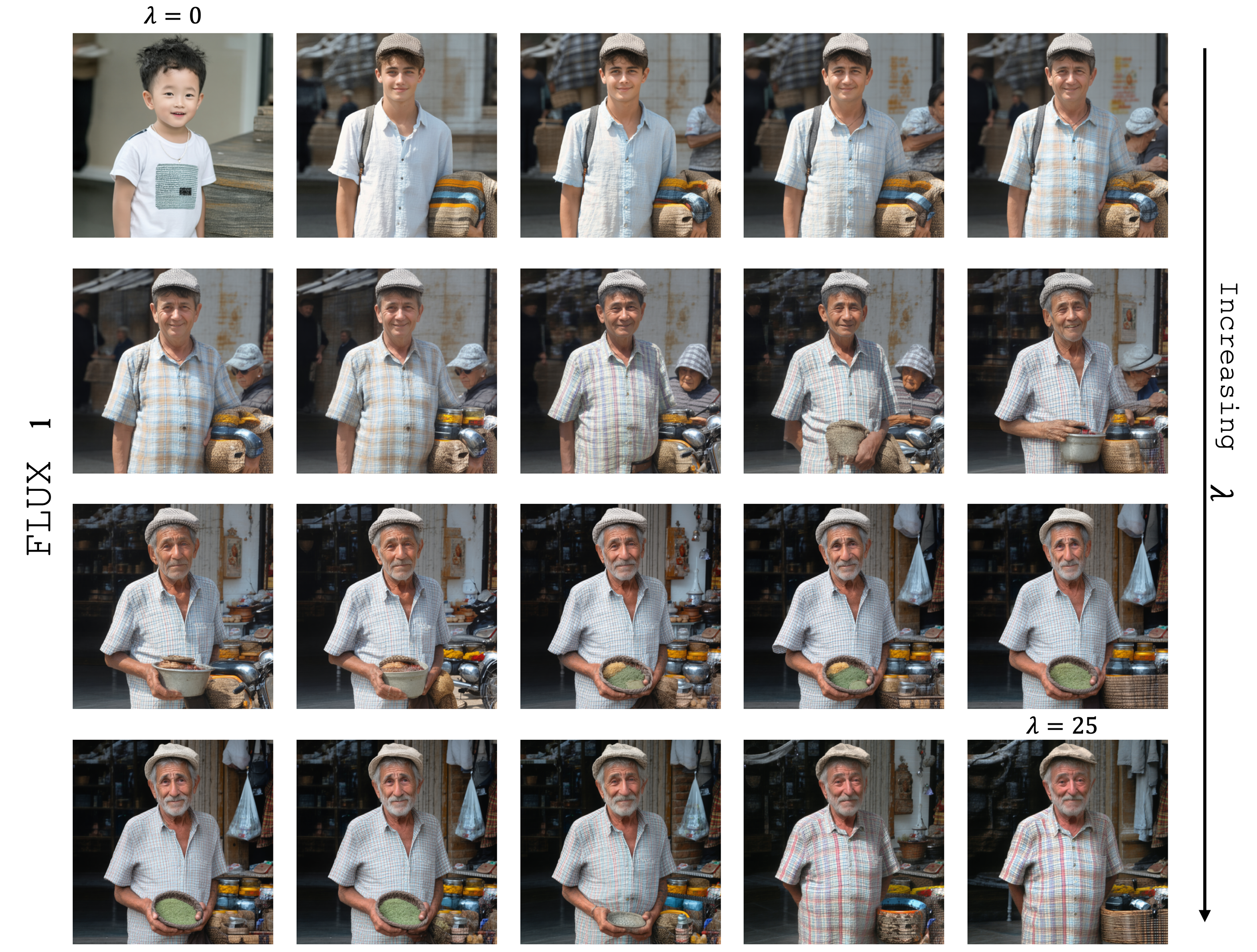}
    \caption{This image shows the effect of the growing value of the $\lambda$ parameter on the generation process. The control aim is to insert the concept of \texttt{'old'} inside the starting picture. Realized on SD3.5}
    \label{fig:lamba-old}
\end{figure}

Figure \ref{fig:lamba-old} illustrates the effect of increasing the steering strength $\lambda$ on the generation process when injecting the concept \texttt{old} into a given starting image. As $\lambda$ grows, the target concept becomes progressively more evident, while low values result in subtle, almost imperceptible changes. This demonstrates the continuous and controllable nature of the proposed concept steering mechanism, allowing fine-grained modulation between preservation of the original content and effective concept insertion.

\begin{figure}
    \centering
    \includegraphics[width=1\linewidth]{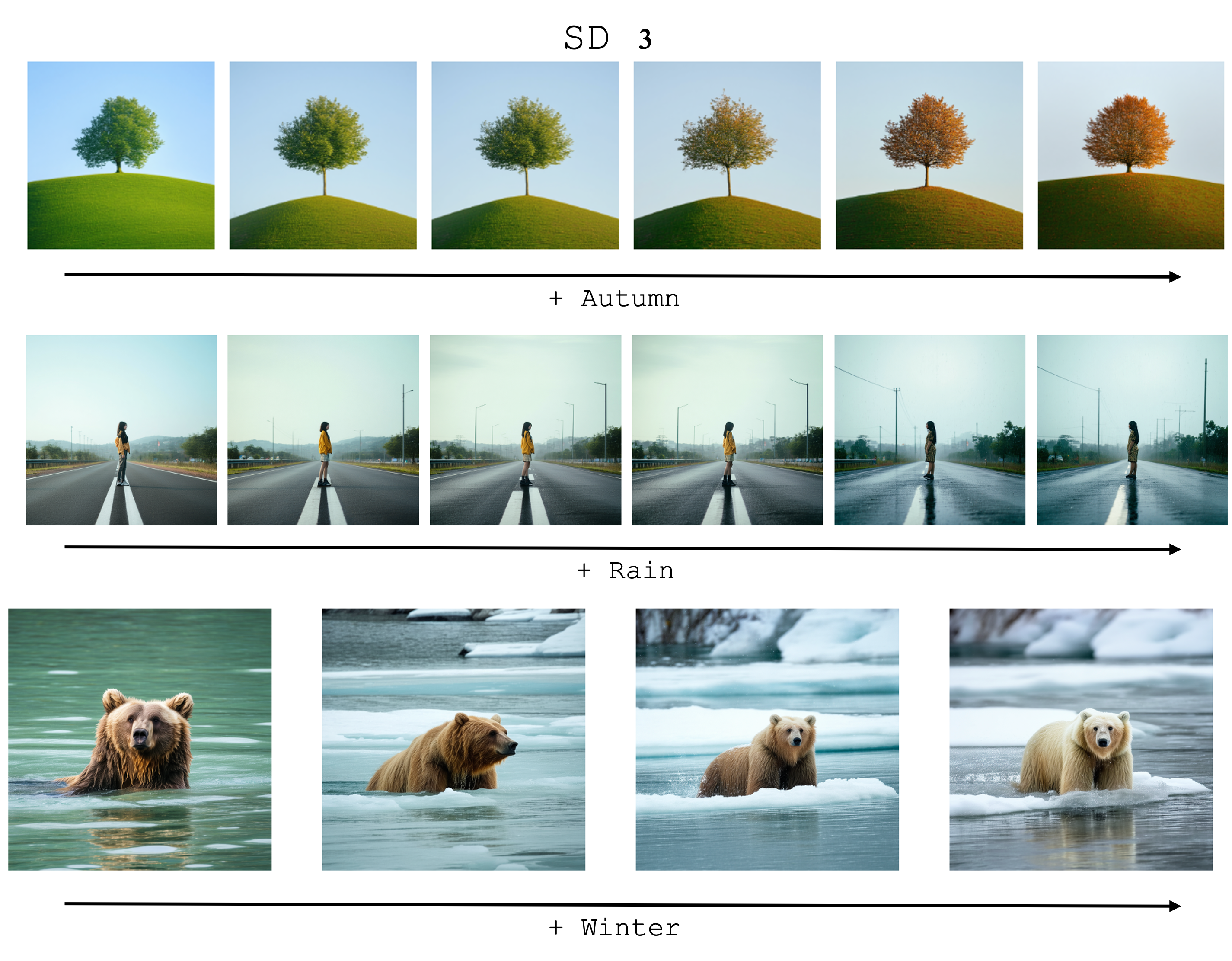}
    \caption{The images illustrate the impact of gradually introducing the steering direction shown in each row.  The concept we want to insert is related to the environment and is subtly introduced while keeping the subject unchanged.}
    \label{fig:season}
\end{figure}

\begin{figure*}[t]
    \centering
    \begin{overpic}[width=1\linewidth]{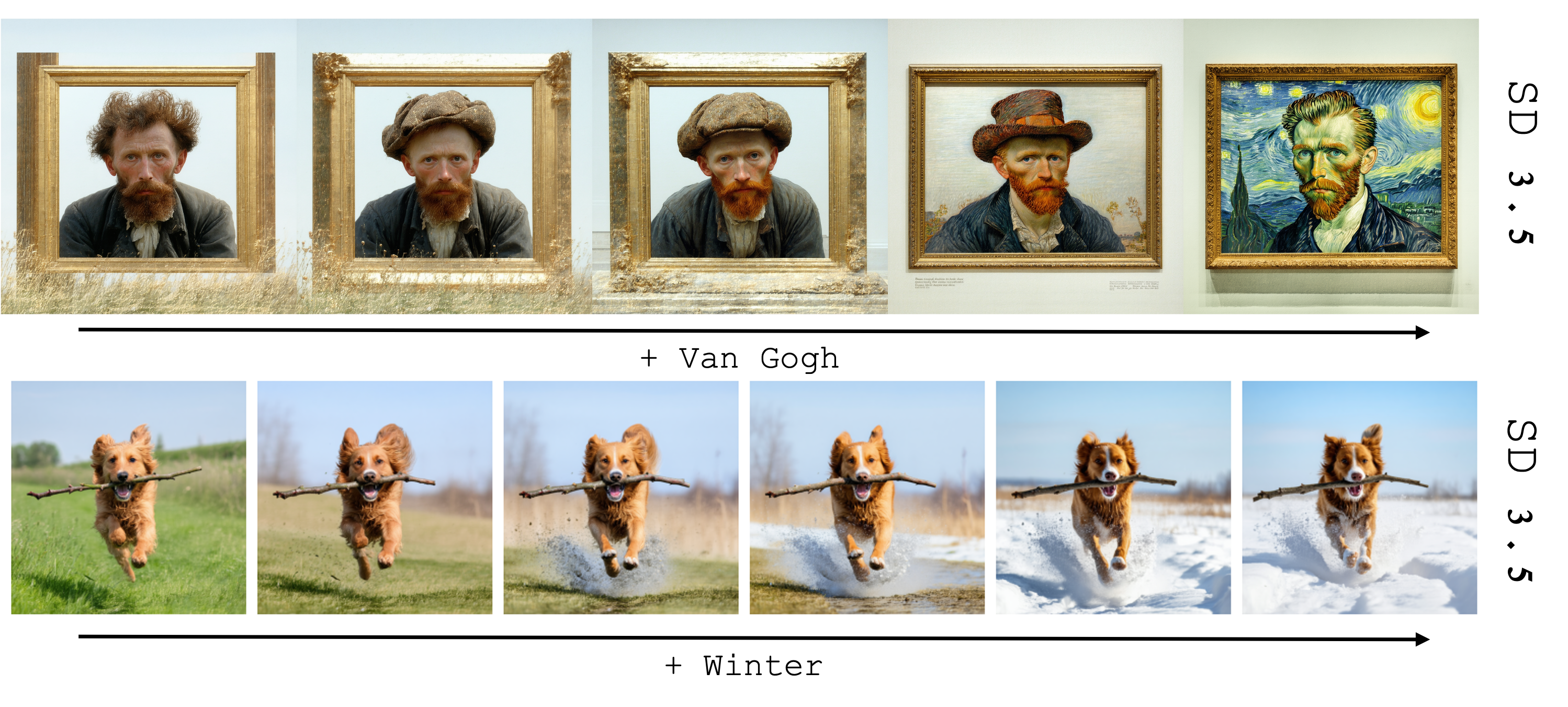}

    \end{overpic}
    \caption{The images illustrate the impact of gradually introducing the steering direction shown in each row.  The concept we want to insert is related to the environment or the style of the subject and is subtly introduced while keeping the subject unchanged.}
    \label{fig:sliders}
\end{figure*}

\begin{figure*}[t]
    \centering
    \includegraphics[width=1\linewidth]{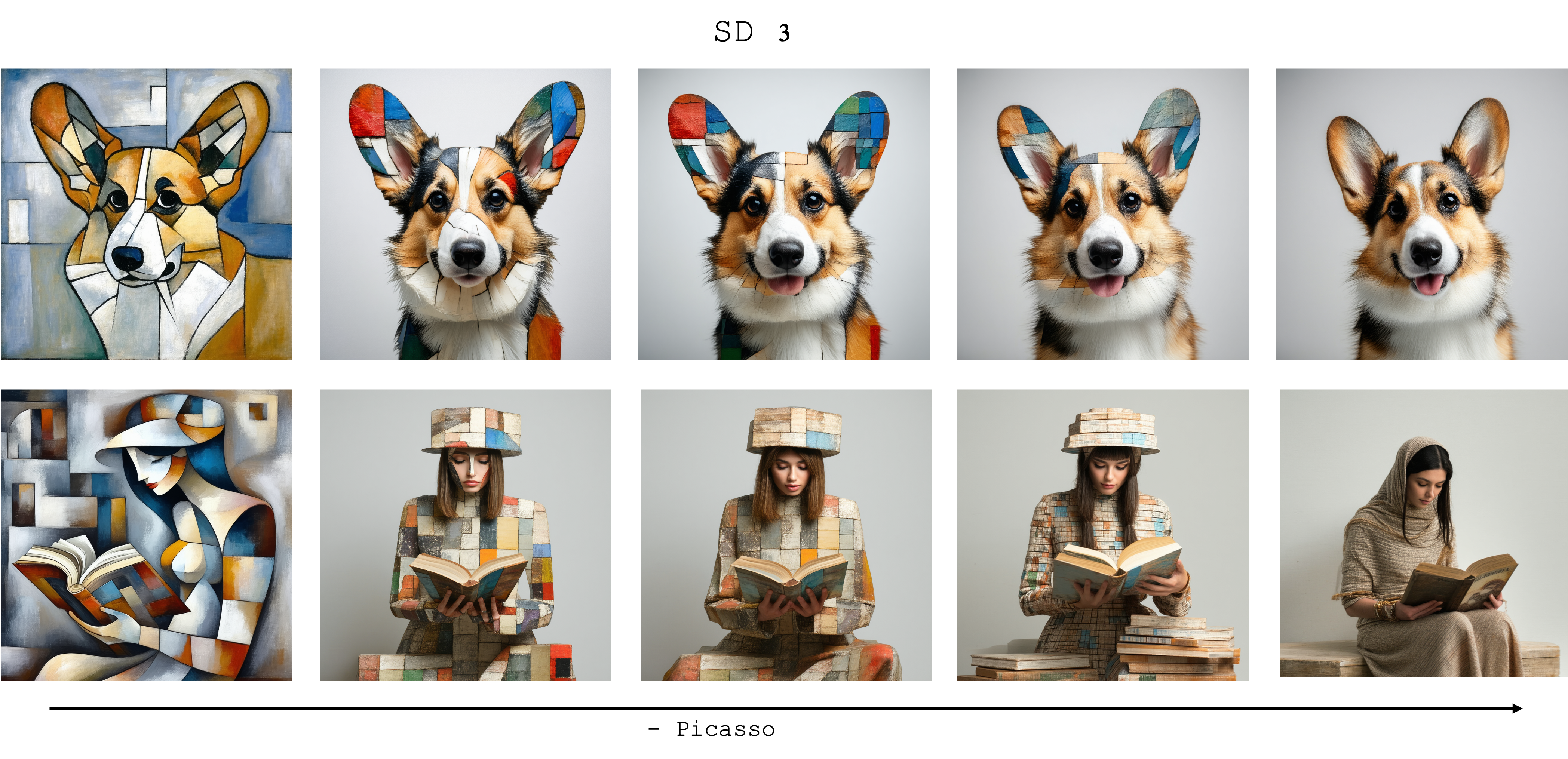}
    \caption{ The images illustrate the impact of gradually introducing the steering direction shown in each row.  The concept we want to remove is related to the environment or the style of the subject and is subtly introduced while keeping the subject unchanged.}
    \label{fig:picasso_dog}
\end{figure*}

Figures \ref{fig:season}, \cref{fig:sliders} and \ref{fig:picasso_dog} focus on environmental and stylistic manipulations. In these examples, the steering direction is gradually introduced across rows, resulting in smooth transitions of background attributes such as season, lighting, or artistic style. Notably, the main subject remains visually stable throughout the process, highlighting the method’s ability to localize semantic changes without inducing unintended alterations to the subject identity or structure.

\begin{figure}
    \centering
    \includegraphics[width=1\linewidth]{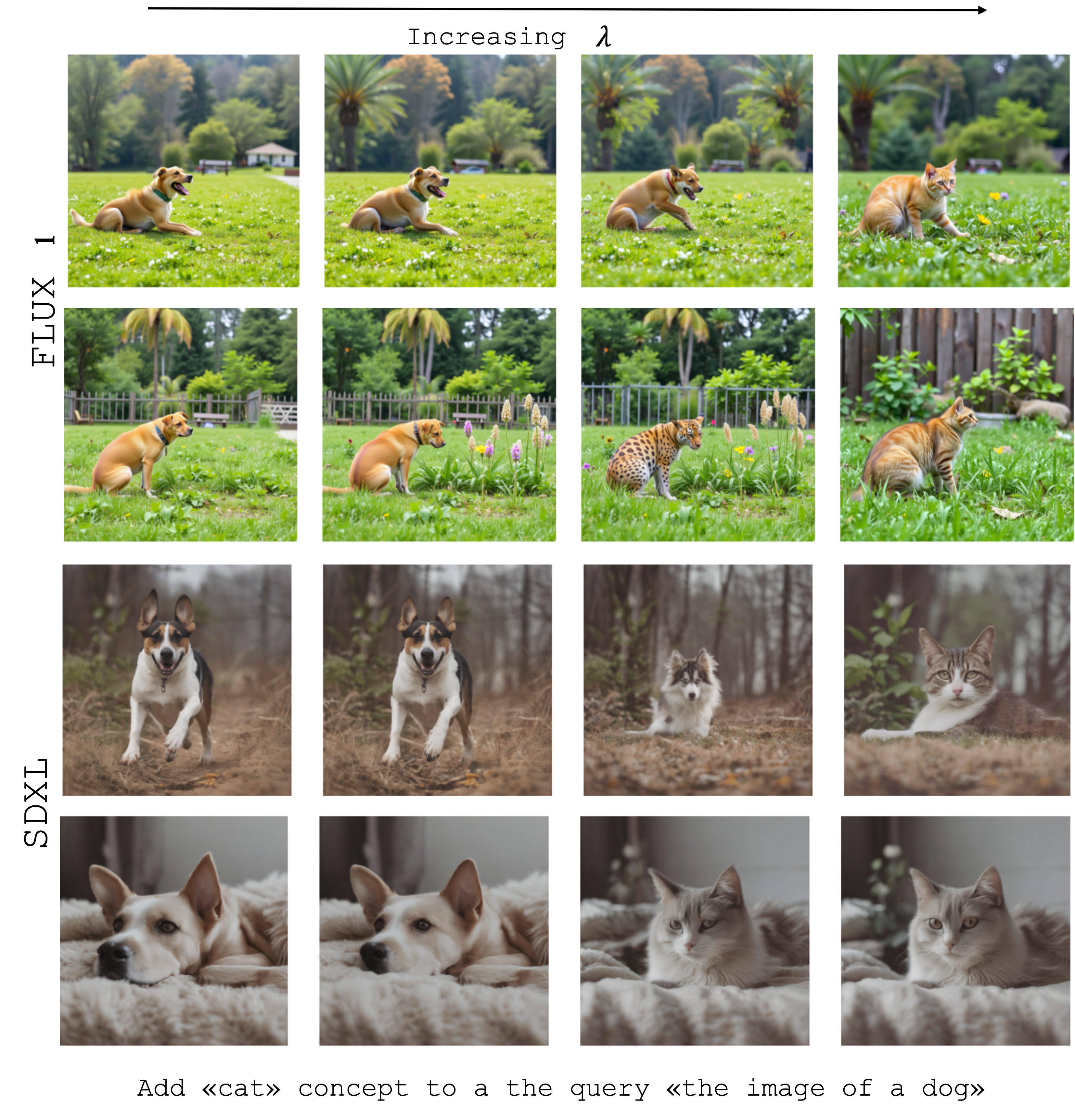}
    \caption{The images illustrate the impact of gradually introducing the steering direction shown in the bottom caption.  In this case, we want to change the subject of our generation into another one; in this case, the steering primarily affects the subject, keeping the environment unchanged.}
    \label{fig:catness}
\end{figure}

In contrast, Figure \ref{fig:catness} showcases a scenario where the steering direction primarily targets the subject itself. Here, the method successfully morphs the subject into a different semantic category while largely preserving the surrounding environment. This behavior underlines the flexibility of the approach, which can selectively affect either subject-centric or context-centric features depending on the chosen steering direction.

\begin{figure}
    \centering
    \includegraphics[width=1\linewidth]{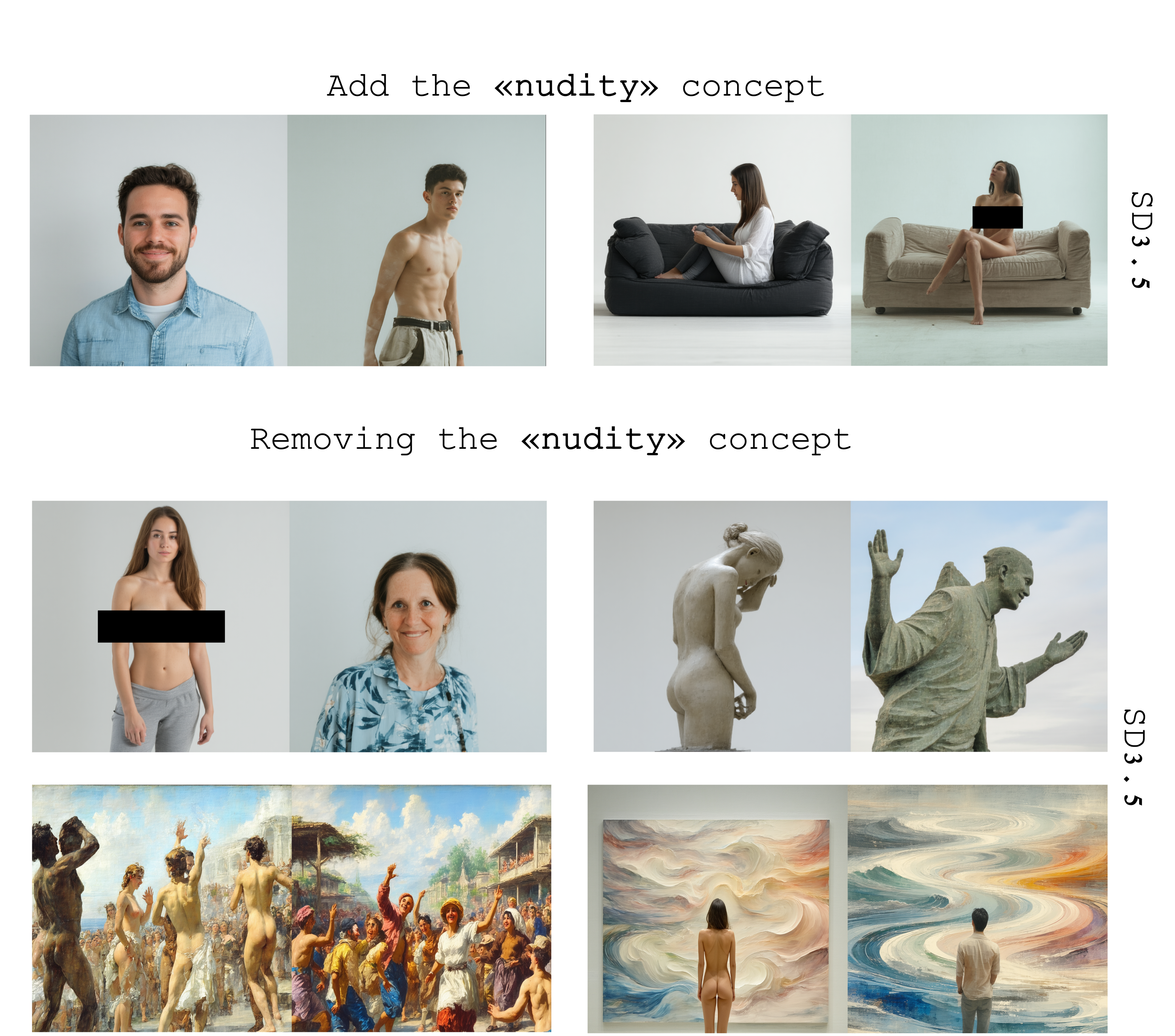}
    \caption{The images show the effect, given a fixed $\lambda=3$, as argued in Sec.~\ref{sec:discussion}, of the removal of the addition of the nudity concept for the SD~3.5 architecture. We can see that the model greatly preserves the context while changing the steered features in both cases.}
    \label{fig:placeholder}
\end{figure}

Figure \ref{fig:placeholder} presents qualitative results for SD 3.5 under a fixed steering strength $\lambda = 3$, following the ablation discussed in Sec.~\ref{sec:abl_prompt_scale}. The examples demonstrate the effect of removing the nudity concept, where the model consistently modifies the targeted attributes while maintaining overall scene coherence and semantic consistency, confirming the robustness of the method across architectural variants.

Finally, Figure \ref{fig:sliders} and Figure \ref{fig:picasso_dog} further illustrate the removal or insertion of stylistic and environmental concepts. Across all cases, changes are introduced progressively and smoothly, without abrupt artifacts, reinforcing the interpretability and controllability of the proposed steering mechanism. Overall, these qualitative results validate the method’s ability to perform precise, disentangled concept manipulation across different models and task settings.



\end{document}